\documentclass{article}

\PassOptionsToPackage{numbers, compress}{natbib}

\usepackage[preprint]{neurips_2025}

\usepackage[utf8]{inputenc} % allow utf-8 input
\usepackage[T1]{fontenc}    % use 8-bit T1 fonts

\usepackage{url}            % simple URL typesetting
\usepackage{booktabs}       % professional-quality tables
\usepackage{amsfonts}       % blackboard math symbols
\usepackage{nicefrac}       % compact symbols for 1/2, etc.
\usepackage{microtype}      % microtypography
\usepackage{graphicx}
\usepackage{amsmath}

\usepackage{makecell}
\usepackage{multirow}
\usepackage{wrapfig}
\usepackage{booktabs}
\usepackage[dvipsnames]{xcolor}
\usepackage[normalem]{ulem}
\usepackage[colorlinks=true, citecolor=Maroon, linkcolor=Maroon]{hyperref}
\usepackage{cleveref}
\useunder{\uline}{\ul}{}

\title{EmbodiedMAE: A Unified 3D Multi-Modal Representation for Robot Manipulation}

\author{Zibin Dong\textsuperscript{$\heartsuit$}, Fei Ni\textsuperscript{$\heartsuit$}, Yifu Yuan\textsuperscript{$\heartsuit$}, Yinchuan Li\textsuperscript{$\diamondsuit$}, Jianye Hao\thanks{Corresponding author: Jianye Hao (jianye.hao@tju.edu.cn).}~~\textsuperscript{$\heartsuit,\diamondsuit$}\\
\textsuperscript{$\heartsuit$}Tianjin University,~~\textsuperscript{$\diamondsuit$}Huawei Noah’s Ark Lab
}

\begin{document}

\maketitle

\begin{abstract}
We present EmbodiedMAE, a unified 3D multi-modal representation for robot manipulation. Current approaches suffer from significant domain gaps between training datasets and robot manipulation tasks, while also lacking model architectures that can effectively incorporate 3D information. To overcome these limitations, we enhance the DROID dataset with high-quality depth maps and point clouds, constructing DROID-3D as a valuable supplement for 3D embodied vision research. Then we develop EmbodiedMAE, a multi-modal masked autoencoder that simultaneously learns representations across RGB, depth, and point cloud modalities through stochastic masking and cross-modal fusion. Trained on DROID-3D, EmbodiedMAE consistently outperforms state-of-the-art vision foundation models (VFMs) in both training efficiency and final performance across 70 simulation tasks and 20 real-world robot manipulation tasks on two robot platforms. The model exhibits strong scaling behavior with size and promotes effective policy learning from 3D inputs. Experimental results establish EmbodiedMAE as a reliable unified 3D multi-modal VFM for embodied AI systems, particularly in precise tabletop manipulation settings where spatial perception is critical.
\end{abstract}

% \begin{figure}[h]
%     \centering
%     \vspace{-10pt}
%     \includegraphics[width=0.8\linewidth]{figures/embodied_mae1.jpg}
%     \caption{Caption}
%     \label{fig:enter-label}
% \end{figure}

\section{Introduction}

Pre-trained Vision Foundation Models (VFMs) have made remarkable progress in visual understanding \citep{caron2021emerging, oquab2024dinov, he2022masked, zhai2023sigmoid, nair2022r3m, majumdar2023vc1, bachmann2022multimae, zhu2025spa}, becoming essential components for embodied AI systems \citep{octo_2023, kim2024openvla, black2024pi0, liu2025rdt, Ze2024DP3, chi2023diffusionpolicy, li2025survey}. As research increasingly demonstrates that 3D spatial understanding can significantly improve robot manipulation capabilities \citep{Ze2024DP3, ke20243ddiffusionactor, li2025pointvla, zhen20243dvla},  the demand for effective 3D VFMs has grown substantially. 3D information provides critical spatial context, enabling robots to accurately localize targets, avoid collisions, and execute complex manipulations. However, despite this increasing need, existing models fall short of meeting requirements.

There are two primary reasons behind the lack of suitable 3D VFMs for embodied AI. \textit{First, a significant domain gap exists in training data}. Mainstream 3D VFMs are trained on outdoor or indoor static scenario datasets \citep{huang2023ponder, zhu2023ponderv2, qian2022pointnext, depth_anything_v1, depth_anything_v2}. These models operate at spatial scales incompatible with tabletop manipulation, where precise understanding within a $20$ cm to $1.5$ m range is crucial and results in a weak understanding of robot-object interactions \citep{Ze2024DP3}. While training 3D embodied-specific VFMs from scratch on robot manipulation datasets seems promising, these efforts are hampered by extremely limited training data \citep{zhu2025spa, qu2025spatialvla}. For example, OpenX Embodiment \citep{vuong2023oxe}, despite being the largest embodied manipulation dataset, contains minimal high-quality 3D information, making it insufficient for effective pre-training. \textit{Second, there is a lack of efficient and scalable model architectures for 3D perception}. Simply integrating 3D information without careful design often degrades robot operation capabilities rather than enhancing them. For example, many advanced 3D VFM architectures demonstrate unexpectedly poor performance in robot manipulation settings, sometimes even underperforming simple MLPs \citep{Ze2024DP3, zhu2024pointcloudmatters}. 

To address these challenges, we propose EmbodiedMAE, a unified 3D multi-modal representation learning framework specifically designed for embodied AI. We first enhance the original DROID dataset \citep{khazatsky2024droid} by extracting high-quality metric depth maps and point clouds for each frame using ZED SDK temporal fusion and AI-augmented enhancement. This creates DROID-3D, a large-scale 3D robot manipulation dataset containing 76K trajectories (350 hours) of high-fidelity interaction data. This dataset provides the scale and quality needed for effective pre-training while maintaining domain compatibility with manipulation tasks. We then develop a multi-modal masked autoencoder that simultaneously learns representations across RGB images, depth maps, and point clouds through stochastic masking and cross-modal fusion. By masking different proportions of each modality and using explicit modal fusion in the decoder, our model learns to infer across modalities, developing powerful spatial perception capabilities and object-level semantic understanding (\Cref{fig:mae_predictions}).

To thoroughly validate our representation model, we conduct extensive evaluations across diverse settings: 40 and 30 simulation tasks from the LIBERO benchmark \citep{liu2023libero} and the MetaWorld benchmark \citep{yu2019metaworld}, 10 real-world tasks on the low-cost open-source SO100 robot platform \citep{cadene2024lerobot}, and 10 tasks on the high-performance xArm robot platform. We use a scaled-down RDT \citep{liu2025rdt} model as the policy backbone to simulate the performance of VFMs in advanced VLA training, and compare EmbodiedMAE against various categories of state-of-the-art (SOTA) VFMs, including vision-centric models, language-augmented models, embodied-specific models, and 3D-aware models. Our experiments demonstrate that EmbodiedMAE consistently outperforms all baseline VFMs in both training efficiency and final performance, exhibits strong scaling behavior with model size, and effectively promotes policy learning from 3D input. These findings establish EmbodiedMAE as a reliable foundation model for embodied AI applications requiring robust 3D visual understanding.

\begin{figure}[t]
    \centering
    \includegraphics[width=1.0\linewidth]{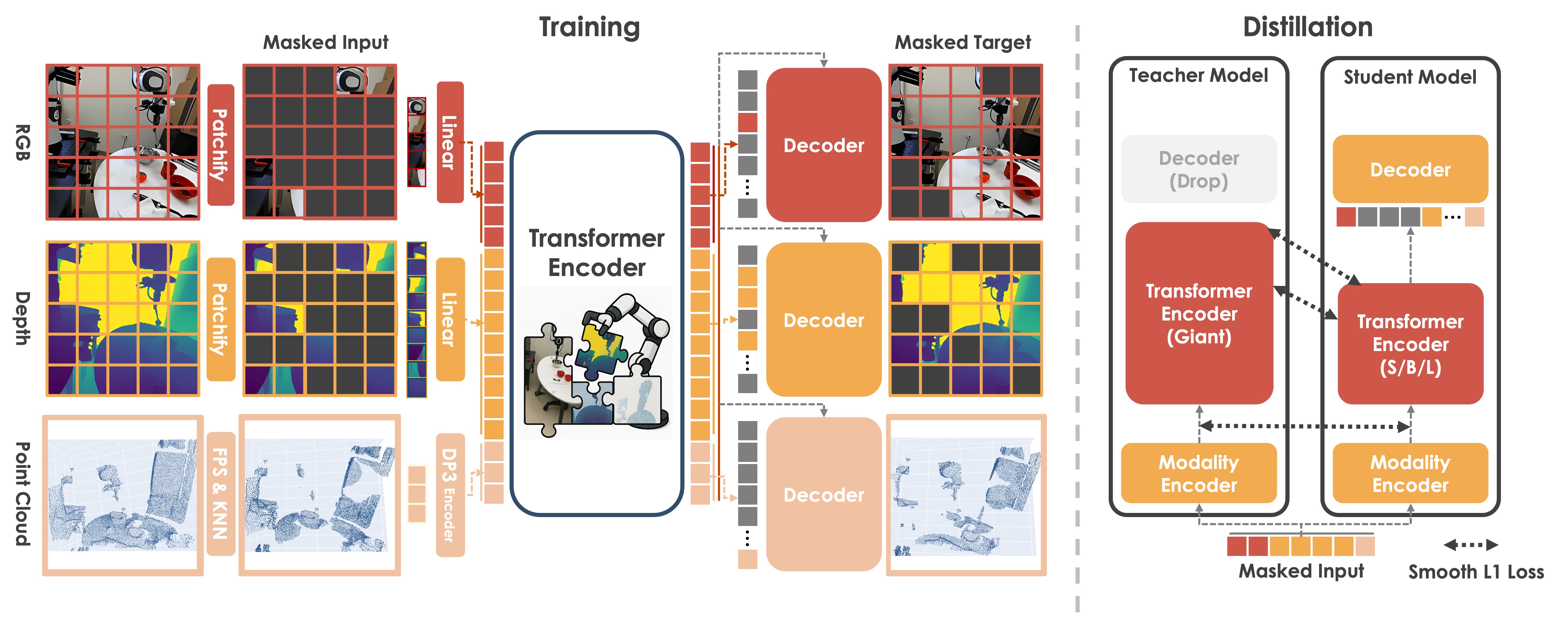}
    \vspace{-20pt}
    \caption{\small{\textbf{Overview of EmbodiedMAE Pre-training.} We pre-train a ViT-Giant scale multi-modal MAE on the large-scale DROID-3D robot manipulation dataset. We fix the total number of unmasked patches across RGB, depth, and point cloud modalities. The mask ratio allocated to each modality is stochastically sampled. After the Giant model pre-training, we distill it to obtain our Small/Base/Large scale models.}}
    \label{fig:main_mae}
    \vspace{-10pt}
\end{figure}

Our contributions can be summarized as follows:
\begin{itemize}
    \vspace{-5pt}
    \item We present EmbodiedMAE, a unified 3D multi-modal representation learning framework for embodied AI that effectively integrates RGB, depth, and point cloud modalities. It achieves SOTA performance in both RGB-only and multi-modal settings while maintaining computational efficiency and scaling properties.

    \item We introduce DROID-3D, a high-quality, large-scale 3D robot manipulation DROID supplement dataset containing 76K trajectories (350 hours) of interaction data with synchronized RGB, depth maps, and point clouds. Unlike previous works that processed only subsets or used low-quality AI-estimated depth, we provide temporally consistent depth by ZED SDK processing, creating a valuable resource for 3D robot learning research.
    
    \item We establish comprehensive evaluation benchmarks for embodied representation learning across diverse settings: simulation tasks from LIBERO and MetaWorld, real-world tasks on a low-cost open-source robot (SO100), and tasks on a high-performance robot (xArm). Our results demonstrate consistent performance improvements across these varied platforms, validating the model's generalization capabilities.
\end{itemize}

\vspace{-5pt}
\section{Methodology}

\subsection{3D Data Collection}

\begin{figure}
    \centering
    \includegraphics[width=1.0\linewidth]{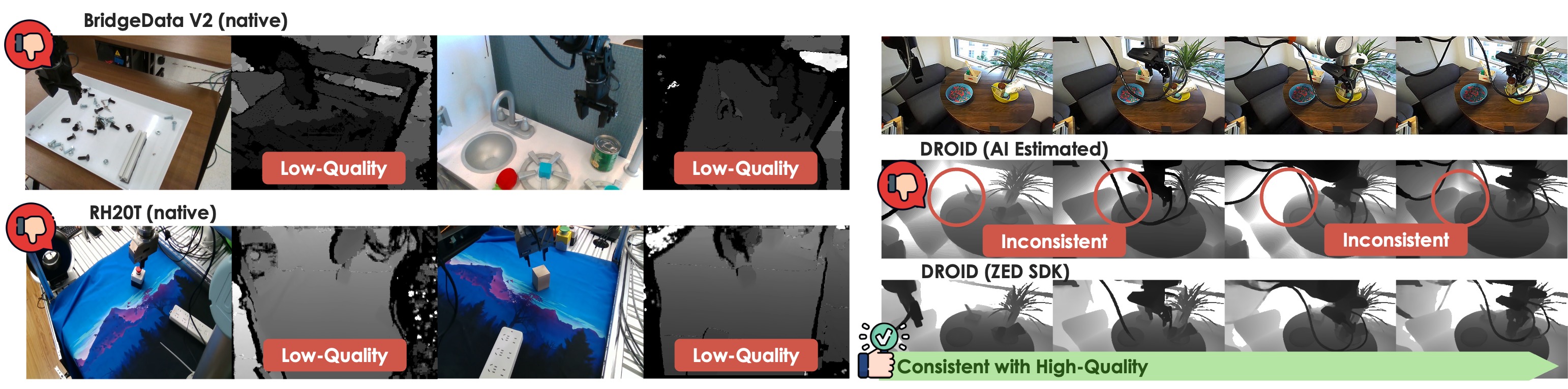}
    \vspace{-15pt}
    \caption{\small{\textbf{Depth Quality Comparison.} We evaluate depth data quality across several mainstream large-scale embodied AI datasets. Both BridgeDataV2 and RH20T exhibit unreliable and noisy depth information. While prior work has explored the use of AI models for depth estimation, we observe that such methods lack temporal consistency. In contrast, our solution, ZED SDK processing, achieves superior and consistent depth quality.}}
    \label{fig:depth_comparison}
    \vspace{-10pt}
\end{figure}

Effective pre-training of our model necessitates a large-scale 3D robot manipulation dataset. We conduct a systematic evaluation of depth data quality across several mainstream large-scale embodied AI datasets, primarily including BridgeDataV2 \citep{walke2023bridgedata}, RH20T \citep{fang2023rh20t}, and DROID \citep{khazatsky2024droid}, as illustrated in \Cref{fig:depth_comparison}. We find significant limitations in existing datasets: BridgeDataV2 contains only 13\% data with 3D information, with available depth maps being of insufficient quality; RH20T exhibits similar issues with unreliable and noisy depth data; while DROID includes stereo image recordings but lacks readily usable 3D annotations. Several previous approaches attempted to address this by estimating depth from 2D images using AI models. For instance, SPA \citep{zhu2025spa} employs CrocoV2-Stereo \citep{Weinzaepfel2023crocov2} to estimate depth for approximately 1/15 of the DROID dataset. We observe that such methods lack precision and temporal consistency, making them unable to accurately capture fine-grained details during robot-object interactions, which are essential for manipulation tasks.

To overcome these challenges, we leverage the fact that the raw DROID dataset preserves ZED camera recordings, which can be processed using ZED SDK to extract high-quality depth information. The ZED SDK integrates multiple techniques that significantly improve depth quality, including temporal fusion to reduce noise and increase consistency, AI-augmented enhancement to refine stereo matching in textureless regions, and hardware-calibrated metric depth to provide accurate absolute distance measurements. Using these high-quality depth maps, we further extract point clouds with the camera's intrinsic matrix. We apply farthest point sampling (FPS) to downsample them to 8,192 points, striking a balance between computational efficiency and geometric fidelity. Unlike SPA's approach of processing only a subset of the DROID dataset, we process the complete collection of 76K trajectories (350 hours of interaction data), requiring nearly 500 hours of processing time. Due to these significant improvements in data quality and coverage, we construct and release DROID-3D as a supplementary resource to the original DROID dataset. We believe it will serve as a valuable resource for pre-training 3D VLA models and foster innovative research in embodied AI, particularly for applications requiring precise spatial understanding for manipulation tasks.

\subsection{Multi-Modal Encoder}

EmbodiedMAE processes three modalities commonly used in robot perception: RGB images, depth maps, and point clouds. Given the robot observation of RGB image $I\in \mathbb R^{3\times H\times W}$, depth $D\in\mathbb R^{1\times H\times W}$, and point cloud $P\in\mathbb R^{M\times 3}$, we first use modal-specific patchifiers to project them into patches $\bar I,\bar D,\bar P\in\mathbb R^{L\times C}$. Then we draw a random binary mask for each modality $m_I,m_D,m_P \in \{0,1\}^L$, and obtain two complementary masked views $I_1=\bar I[m_I], I_2=\bar I[1-m_I]$, similar for $D$ and $P$. We use a Vision Transformer (ViT) $f$ to process the unmasked patches and obtain the joint representation $h = f(I_1, D_1, P_1)$. 

\textbf{Masking Strategies.} Effective masked autoencoding requires masking a large portion of input tokens during training, and the specific masking strategy has a significant impact on learned representations \citep{bachmann2022multimae, he2022masked}. Following \citet{bachmann2022multimae}, we fix the total number of unmasked patches across all modalities, i.e., the number of ones in $(m_I,m_D,m_P)$ is fixed, and allocate them according to a symmetric Dirichlet distribution: $(\lambda_I, \lambda_D, \lambda_P) \sim \text{Dir}(\alpha),$ where $\lambda_I + \lambda_D + \lambda_P = 1$ and each $\lambda \geq 0$. The concentration parameter $\alpha$ controls the diversity of masking proportions. When $\alpha = 1$, the distribution is uniform over the simplex, assigning equal likelihood to all valid combinations. Lower values ($\alpha \ll 1$) tend to concentrate sampling on a single modality, while higher values ($\alpha \gg 1$) produce more balanced allocations across modalities. We intentionally avoid introducing any modality bias by keeping the distribution symmetric, aiming to maintain flexibility for a variety of downstream tasks and input configurations.

\textbf{Modal Patchifiers.} For RGB and depth maps, we break them into 16$\times$16-size patches, i.e., $L = \frac{H \cdot W}{16^2}$, and we incorporate 2D sine-cosine positional embeddings after a linear projection \citep{dosovitskiy2021vit, hugo2021deit}. For point clouds, we apply Farthest Point Sampling (FPS) to select $N$ cluster centers, and then use K-Nearest Neighbors (KNN) to group each center with its $K$ nearest neighbors, forming $N$ point groups of $K+1$ points each, i.e., $L=N$. Each group is normalized and encoded using a DP3 encoder \citep{Ze2024DP3} to generate token embeddings, while each group center is processed by an MLP to create positional embeddings \citep{pang2022pointmae}. We omit explicit modality-type embeddings, as the bias term in each projection layer implicitly encodes modality-specific information. These tokens are masked, concatenated, and passed to the ViT encoder to produce the joint representations.

\textbf{Transformer Encoder.} We implement the same ViT structure as DINOv2 \citep{oquab2024dinov}, with the exception of removing the \texttt{[CLS]} token. This design choice allows us to initialize the ViT directly from DINOv2 pre-trained weights, thereby enhancing its general capabilities.

\begin{figure}[t]
    \centering
    \makebox[\textwidth][c]{\includegraphics[width=1.0\linewidth]{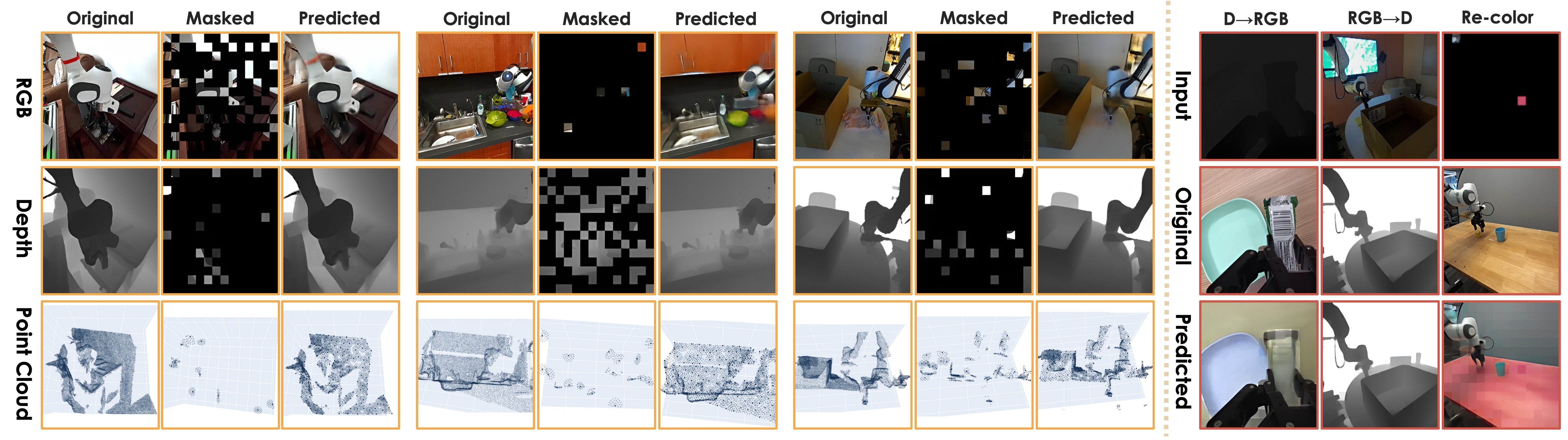}}
    \vspace{-15pt}
    \caption{\small{\textbf{EmbodiedMAE Visual Predictions.} We evaluate its visual predictions under three settings: \textbf{(a)} Two modalities are almost masked, leaving one modality as the major infer source (column 1-9). \textbf{(b)} Model predicts one modality from another one (column 10-11). \textbf{(c)} Model is allowed to see a modified RGB patch during depth-to-RGB prediction, where the color of the visible patch is altered (column 12).}}
    \label{fig:mae_predictions}
    \vspace{-11pt}
\end{figure}

\vspace{-5pt}
\subsection{Multi-Modal Decoder}

The decoder is only used during EmbodiedMAE training, where it reconstructs the masked portions of each modality based on the visible tokens and learned \texttt{[MASK]} tokens. 

Specifically, the decoder employs cross-attention to enable explicit fusion across modalities. Visible tokens from each modality are projected and concatenated with \texttt{[MASK]} tokens, then augmented with positional embeddings to form the query sequence. Meanwhile, all visible patches are projected and enhanced with modality encodings to form the key and value sequences. The fused features are then fed into a smaller, modality-shared ViT decoder to produce the final hidden states. Modality-specific MLP heads generate the reconstruction outputs: masked RGB and depth patches, and normalized point coordinates for point cloud groups. Suppose that $(h_I, h_D,h_P)=f(I_1,D_1,P_1)$ are modality representations, the decoder outputs can be expressed as $g_I(h_I, h)$, $g_D(h_D, h)$, and $g_P(h_P, h)$, corresponding to each modality. Notably, our design shares transformer components across modalities, reducing computational cost by approximately a factor of three. We adopt a simple mean square error (MSE) loss:
\begin{equation}
    \mathcal L_{\text{MAE}}=\mathbb E_{(I,D,P)\sim\mathcal D,\text{Dir}(\alpha)}\left[\underbrace{\Vert g(h_I,h)-I_2 \Vert^2}_{\text{\tiny {RGB}}}+\underbrace{\Vert g(h_D,h)-D_2 \Vert^2}_{\text{\tiny Depth}}+\underbrace{\Vert g(h_P,h)-P_2 \Vert^2}_{\text{\tiny PointCloud}}\right],
\end{equation}
where the decoder outputs $g_I(h_I, h)$ and $g_D(h_D, h)$ are $l_2$-normalized, while $g_P(h_P, h)$ is group center-normalized, following MAE's finding that normalized targets yield better performance. \citep{he2022masked}.

\subsection{Model Distillation}
\label{sec:distill}
Following \citet{oquab2024dinov}, we first train a ViT-Giant EmbodiedMAE model from scratch on the DROID-3D dataset, then distill it into Small, Base, and Large variants. Both teacher and student models receive identical masked inputs $(I_1,D_1,P_1)$, with the teacher model kept entirely frozen. Rather than simply copy the final outputs, we apply feature-level supervision at strategically selected network depths to ensure comprehensive knowledge transfer. Specifically, we align features at three critical positions in the network hierarchy: (Bottom) immediately after the modal patchifiers to capture low-level perceptual features, (Top) at the final hidden layer to preserve high-level semantic understanding, and (Middle) at a middle layer positioned at 3/4 of the encoder depth to transfer intermediate representations \citep{bai2022dmae} (For example, when distilling from a 24-layer ViT-L teacher to a 12-layer ViT-B student, the 9th layer of the student aligns with the 18th layer of the teacher.). We adopt trainable linear projections before computing alignment losses to accommodate dimensional differences between teacher and student features. Formally, we denote the feature alignment pairs $(y^j, h^j) \in A$, where $y^j$ and $h^j$ represent the $j$-th pair of hidden states from teacher and student models, respectively, and $l^j$ is the linear projector. The feature alignment loss can be expressed as:
\begin{equation}
    \mathcal{L}_{\text{Align}}=\sum\nolimits_{(y^j,h^j)\in A}\text{SmoothL1}\left(y^j,l^j(h^j)\right).
\end{equation}
We train student models by jointly optimizing the standard multi-modal MAE reconstruction loss and the feature alignment loss (\Cref{fig:main_mae}, Distillation part):
\begin{equation}
\mathcal{L}_{\text{Distill}}=\mathcal{L}_{\text{MAE}}+\beta\cdot\mathcal{L}_{\text{Align}},
\end{equation}
where $\beta>0$ controls the balance between mask autoencoding and feature alignment. This approach enables our smaller models to achieve performance closer to the Giant model while maintaining computational efficiency, making them practical in resource-constrained robotics applications.

\subsection{Put All Together}

\begin{wrapfigure}{r}{0.5\textwidth}
\centering
    \vspace{-10pt}
    \includegraphics[width=0.5\textwidth]{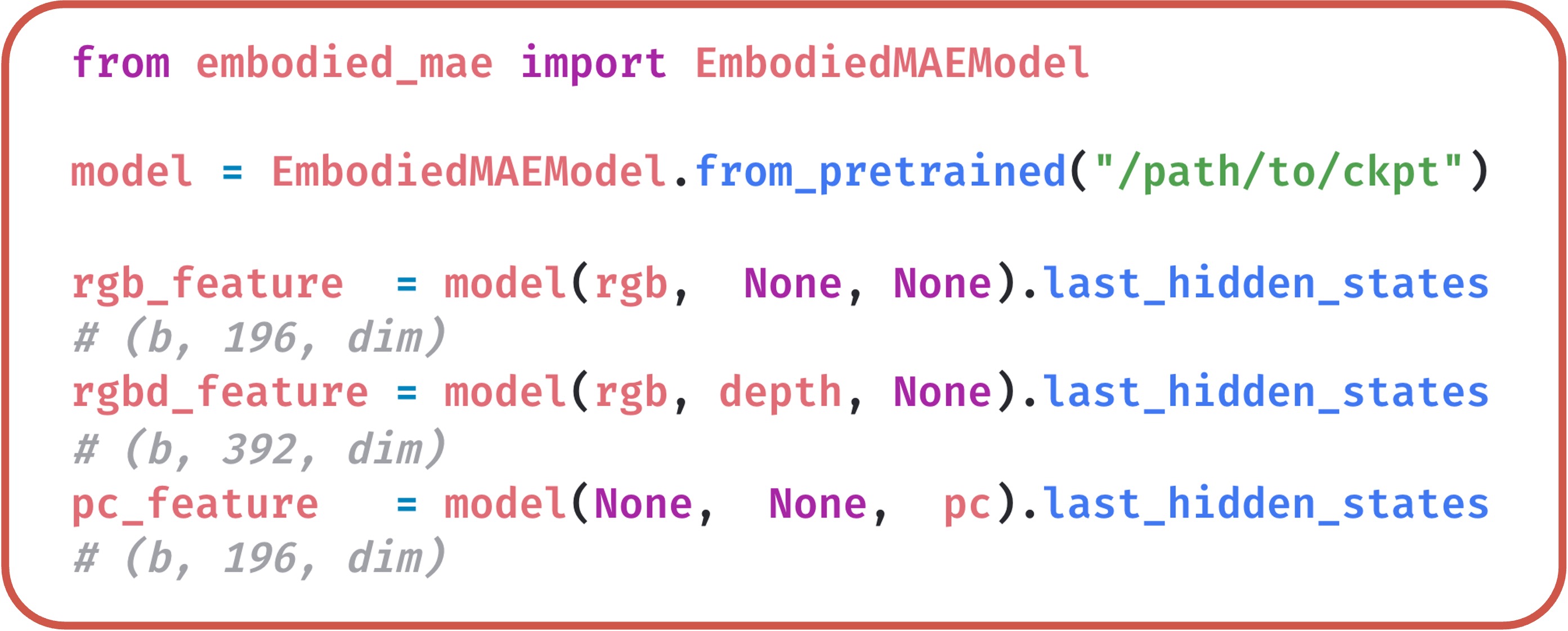}
    \caption{\small{\textbf{Usage Example.} We follow the Huggingface Transformers convention to make EmbodiedMAE highly user-friendly and easy to integrate.}}
    \label{fig:usage_example}
    \vspace{-10pt}
\end{wrapfigure}

Building on our architectural design described above, we first pre-train the Giant-scale model and subsequently distill it into more computationally efficient Small, Base, and Large variants on the DROID-3D dataset. We employ AdamW optimizer with a weight decay of 0.01. The base learning rate is set at 1.5e-4, incorporating an initial warmup period followed by a cosine schedule decay. We apply a 0.1 gradient norm clip to stabilize training. All computational workflows utilize \texttt{bfloat16} precision, which substantially reduces memory requirements and computational costs while maintaining numerical stability. During the pre-training phase, we maintain 96 unmasked patches across all modalities, representing approximately 1/6 of the total patch count. For the distillation phase, we further reduce the number of unmasked patches to 60, approximately 1/10 of the total. This extremely aggressive masking approach significantly decreases training costs without compromising representational quality, as the student models benefit from the teacher's already robust understanding of multi-modal relationships. The efficient training strategy enables us to complete Giant model pre-training on 8 NVIDIA L40 48G GPUs and the distillation phases on 4 NVIDIA GeForce RTX 4090 24G GPUs. We provide hyperparameter configuration list in \Cref{tab:hyperparams}. 

Our codebase follows the Huggingface Transformers \citep{wolf-etal-2020-transformers} convention, making EmbodiedMAE highly user-friendly. It ensures that researchers can easily incorporate our models into existing robotics pipelines with minimal adaptation effort. A simple usage example is illustrated in \Cref{fig:usage_example}.

\section{Experiments}

In this section, we present evaluation results of EmbodiedMAE across both simulation and real-world robotic manipulation tasks. Our experiments are designed to address three key research questions:

\textbf{(RQ1)} Does EmbodiedMAE learn features that integrate information across different modalities?

\textbf{(RQ2)} How does EmbodiedMAE perform compared to SOTA VFMs in robot manipulation tasks?

\textbf{(RQ3)} Can EmbodiedMAE enable efficient robot learning in real-world environments for both low-cost and high-performance robot platforms?

\vspace{-5pt}
\subsection{Experimental Setup}

\begin{wrapfigure}{tbpr}{0.4\textwidth}
\centering
    \vspace{-20pt}
    \includegraphics[width=0.4\textwidth]{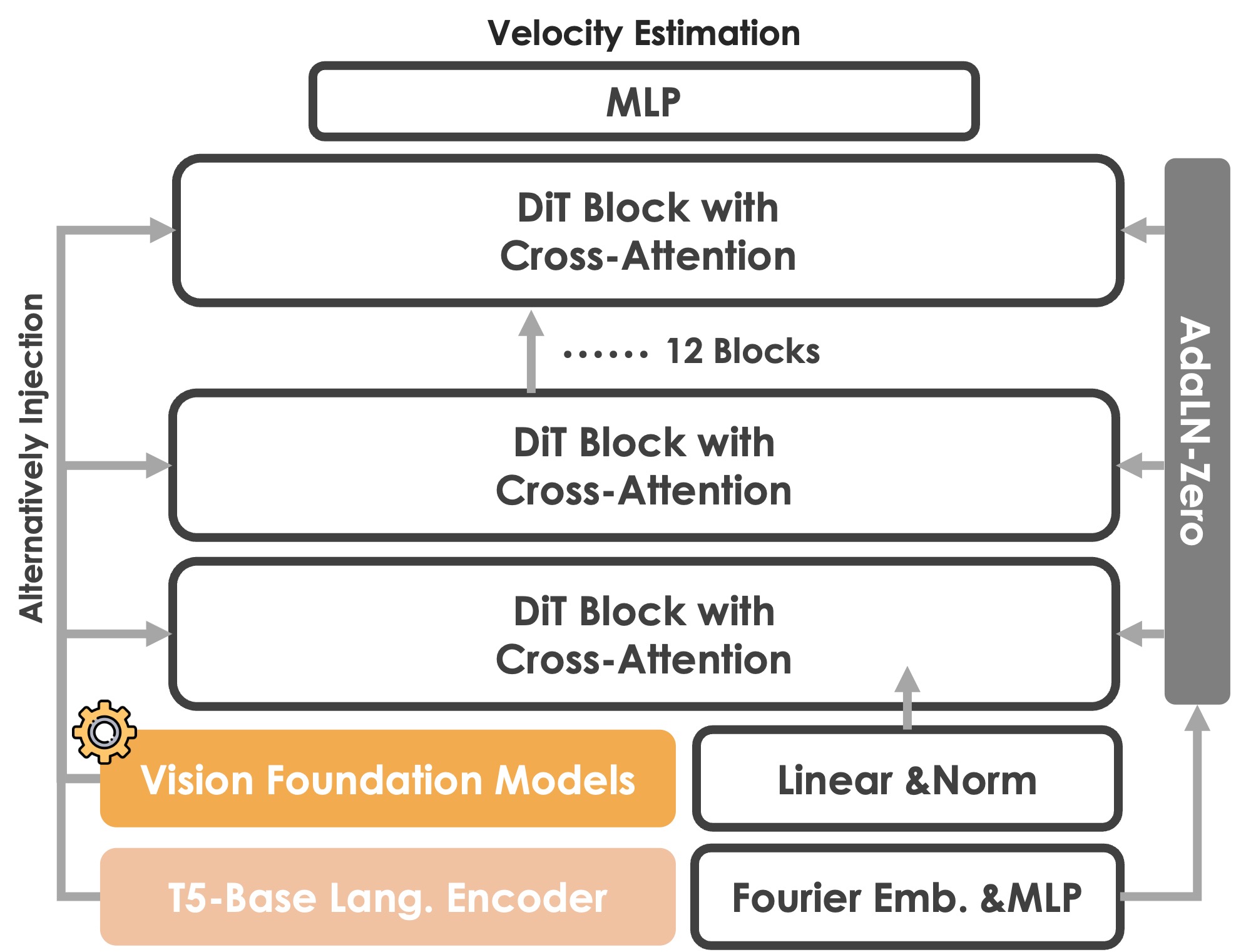}
    \vspace{-15pt}
    \caption{\small{\textbf{Policy Network for All VFMs.} We adopt a compact RDT as the policy network, in which only VFMs are modular.}}
    \label{fig:policy_backbone}
    \vspace{-15pt}
\end{wrapfigure}

\textbf{Policy Network.} To evaluate how effectively different VFMs support advanced VLA models, we adopt a compact RDT \citep{liu2025rdt} (approximately 40M parameters) as our policy network. This architecture has demonstrated excellent scalability and strong performance in diffusion-based policy learning. As shown in \Cref{fig:policy_backbone}, all baselines and EmbodiedMAE share the same architecture, ensuring fair comparison by isolating the visual representation component. See \Cref{appendix:policy_network} for more policy network details.

\textbf{Baselines.} To enable a comprehensive comparison, we benchmark against several SOTA VFMs with diverse design principles: DINOv2-Large \citep{oquab2024dinov} (vision-centric), SigLIP-Large \citep{zhai2023sigmoid} (language-contrastive), R3M-Resnet50 \citep{nair2022r3m}, VC-1 \citep{majumdar2023vc1}, and SPA \citep{zhu2025spa} (embodied-specific). Notably, SPA incorporates implicit 3D spatial priors during training, making it particularly relevant for comparison with our multi-modal approach.

\textbf{Benchmarks.} Our simulation evaluations are based on the LIBERO and MetaWorld benchmarks. LIBERO includes 40 tasks in four task suites: \textit{Goal}, \textit{Spatial}, \textit{Object}, and \textit{Long}. MetaWorld includes 30 tasks from various difficulty levels. For real-world experiments, we deploy the models on two robot platforms: The SO100 robot (low-cost, open-sourced, equipped with dual RGB cameras) evaluated on 10 tasks in suites: \textit{Pick\&Place}, \textit{MoveTo}, \textit{Wipe}, and \textit{Unfold}; The xArm robot (higher-precision, equipped with one Intel RealSense L515 LiDAR camera) evaluated on 10 tasks in suites: \textit{Pick\&Place}, \textit{Pot}, \textit{Pour}, and \textit{Moka}. We show detailed task configurations in \Cref{appendix:benchmarks}.

\vspace{-5pt}
\subsection{MAE Predictions (RQ1)}

To assess the ability of EmbodiedMAE to integrate information across modalities, we design a series of controlled experiments probing its cross-modal fusion capabilities. Our evaluation focuses on three settings: \textcolor{Maroon}{(a) Extreme modality inference}: We mask most patches from two modalities, leaving primarily one modality as the inference source (\Cref{fig:mae_predictions}, columns 1-9). \textcolor{Maroon}{(b) Cross-modal translation}: We test the model's ability to predict one entire modality from another, specifically RGB from depth (column 10) and depth from RGB (column 11). \textcolor{Maroon}{(c) Re-coloring}: We allow the model to see a deliberately altered RGB patch during depth-to-RGB prediction (column 12), where the color of the visible patch is modified to assess semantic understanding. Our results demonstrate that EmbodiedMAE effectively leverages available modalities to reconstruct missing information, suggesting strong cross-modal alignment. In column 10, the predicted RGB from depth lacks precise color information but maintains structural fidelity, indicating the model has learned to separate geometric and appearance features. Similarly, in column 11, depth predictions from RGB show smoothed object boundaries compared to ground truth, revealing a learned prior for depth continuity. Most notably, in the re-coloring setting (column 12), when injecting an altered RGB patch during depth-to-RGB reconstruction, only the corresponding object (table) adopts the modified color while surrounding elements (background, robot, cup) maintain their original appearance. This suggests EmbodiedMAE has implicitly learned object-level semantic segmentation and can propagate semantic information based on contextual cues, despite never being explicitly trained for segmentation.

These visualizations collectively demonstrate that EmbodiedMAE possesses strong multi-modal fusion capabilities, enabling it to enhance spatial understanding in 3D embodied perception tasks.

\begin{figure}
    \centering
    \makebox[\textwidth][c]{
    \includegraphics[width=1.0\linewidth]{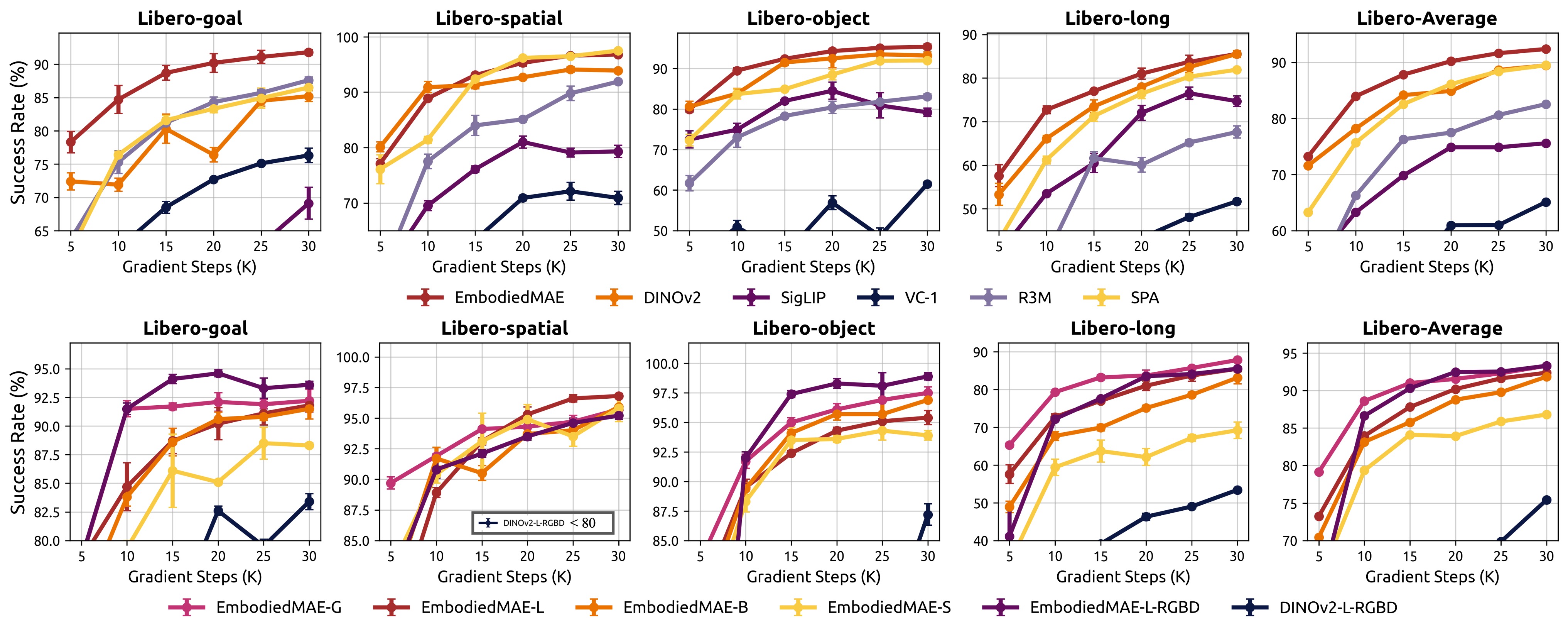}}
    \vspace{-20pt}
    \caption{\small{\textbf{Learning curve on LIBERO benchmark.} Each task is evaluated across 150 trials. Our model surpasses all baselines on the LIBERO benchmark and demonstrates scaling capabilities, with performance increasing proportionally with model size. Our model effectively leverages 3D information to further enhance policy performance, whereas naively incorporating depth information results in performance degradation.}}
    \label{fig:libero_main}
    \vspace{-5pt}
\end{figure}

\vspace{-5pt}
\subsection{Overall Comparison (RQ2)}

\begin{table}[]
\centering
\label{tab:metaworld_scores}
\vspace{-0pt}
\caption{\small{\textbf{Success rate on MetaWorld benchmark.} We report the average success rate for each difficulty level. The numerical suffix following each level indicates the number of tested tasks. Note that \textbf{Average} row represents the average across all tasks rather than the three difficulty levels. Highest scores are emphasized with bold.}}
\scalebox{0.9}{
\begin{tabular}{l|ccccc|cc}
\toprule
MetaWorld  & R3M  & SigLIP & DINOv2 & SPA        & EmbodiedMAE & \multicolumn{1}{c}{DINOv2} & \multicolumn{1}{c}{EmbodiedMAE} \\
\scriptsize{Difficulty Level}  &      &        &        &            &             & \multicolumn{1}{c}{\scriptsize{-RGBD}} & \multicolumn{1}{c}{\scriptsize{-RGBD}} \\
\midrule
Easy (18)      & 74.1 & 76.4   & 79.8   & 80.9       & {\ul 81.8}  & 61.9          & \textbf{85.2}    \\
Medium (9)    & 28.1 & 32.7   & 57.1   & {\ul 62.8} & 60.4        & 35.6          & \textbf{63.2}    \\
Very Hard (3) & 49.8 & 14.0   & 56.4   & 55.8       & 57.8        & \textbf{65.6} & {\ul 61.6}       \\
\midrule
\textbf{Average}      & 57.9 & 57.0   & 70.7   & {\ul 73.0} & {\ul 73.0}  & 54.4          & \textbf{76.2}   \\
\bottomrule
\end{tabular}}
\vspace{-10pt}
\end{table}

In this section, we evaluate SOTA VFM baselines, EmbodiedMAE, and several its variants (in terms of model scale and input modality) on the LIBERO and MetaWorld benchmark. We report learning curves on LIBERO in \Cref{fig:libero_main} and success rate on MetaWorld in \Cref{tab:metaworld_scores}. \textit{Unless otherwise specified, ``EmbodiedMAE'' refers to the Large-scale, RGB-only variant.}

% EmbodiedMAE outperforms all baselines on the LIBERO benchmark and shows comparable performance to SPA on the MetaWorld benchmark. 
\textcolor{Maroon}{\textbf{\textit{Finding 1:}}} \textbf{EmbodiedMAE consistently outperforms all baseline VFMs in terms of both training efficiency and final performance.} Among the baselines, SPA and DINOv2 are the most competitive ones. SPA shows score gains on tasks where spatial understanding is crucial, e.g., LIBERO-Spatial and MetaWolrd, and performs comparably to DINOv2. The language-contrastive model, SigLIP, performs poorly across all embodied tasks, consistent with findings from \citet{zhu2025spa}. R3M and VC-1, although specifically designed for robot learning, do not demonstrate clear advantages.

\textcolor{Maroon}{\textbf{\textit{Finding 2:}}} \textbf{EmbodiedMAE exhibits strong scaling behavior with model size.} Performance improves monotonically as model capacity increases. Among all the variants, only the Small variant shows unstable performance on LIBERO-Goal and LIBERO-Object suites. The Base and Large models achieve similar performances, with the Large model slightly ahead. The Giant model consistently delivers superior performance, particularly in training efficiency. These results suggest EmbodiedMAE to be an effective training paradigm for scaling multi-modal representation learning.

\textcolor{Maroon}{\textbf{\textit{Finding 3:}}} \textbf{EmbodiedMAE promotes policy learning from 3D input.} When provided with RGBD inputs, EmbodiedMAE establishes a substantial performance gap over other baselines on both LIBERO and MetaWorld benchmarks. Remarkably, our Large-scale RGBD model even outperforms the Giant-scale RGB-only model on LIBERO-Goal and LIBERO-Object suites, and performs comparably on average across the LIBERO benchmark. In contrast, adding a trainable depth branch for DINOv2 (See \Cref{appendix:dino_rgbd} for details of this variant) can degrade performance relative to RGB-only input, consistent with observations in \citet{zhu2024pointcloudmatters}. These findings establish EmbodiedMAE as a reliable VFM for scenarios requiring 3D visual understanding.

\vspace{-5pt}
\subsection{Real-World Experiments (RQ3)}

\begin{figure}[t]
    \centering
    \includegraphics[width=1.0\linewidth]{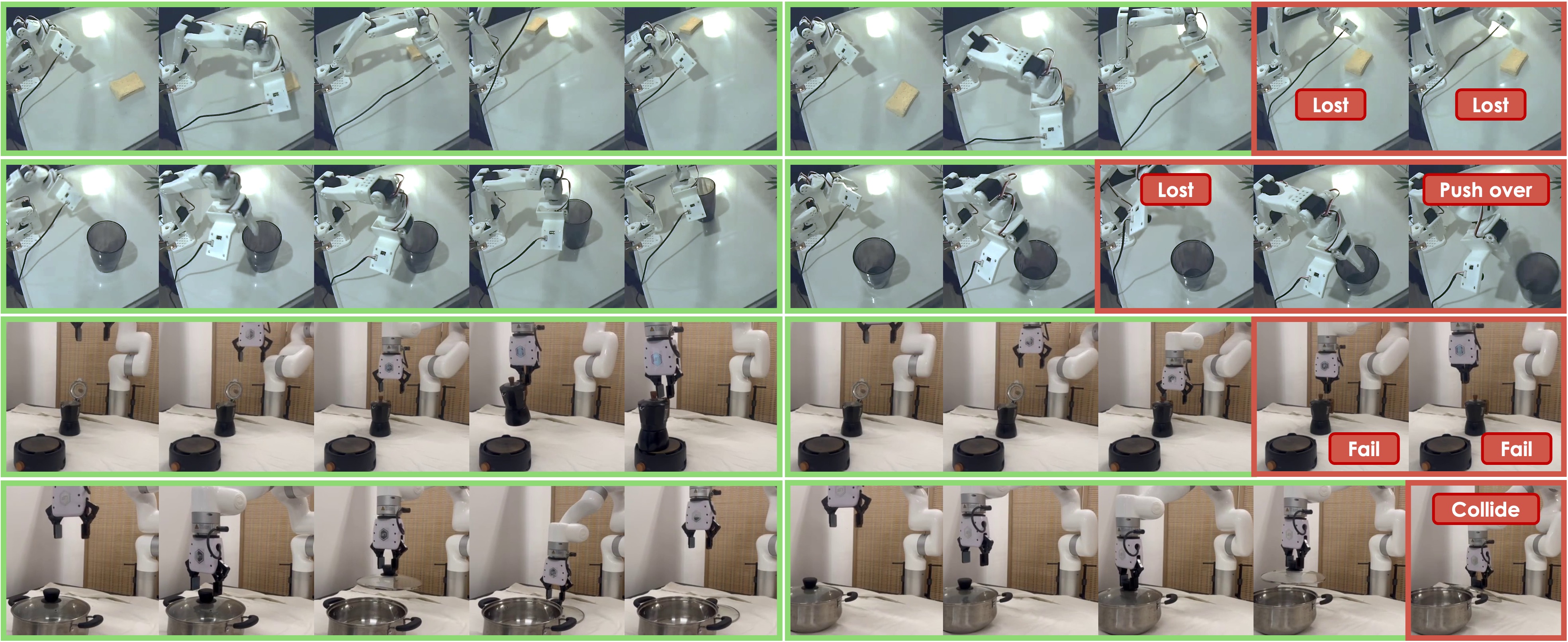}
    \caption{\small{\textbf{Successful rollouts of EmbodiedMAE (Left) and typical failure cases of baselines (Right).}  Baseline models often fail due to inaccurate localization, leading to object loss, grasp failure, or collisions. In contrast, EmbodiedMAE benefits from stronger spatial perception and avoids such errors more effectively.}}
    \label{fig:real_robot_demos}
    \vspace{-5pt}
\end{figure}

% \begin{figure}[t]
%     \centering
%     \includegraphics[width=1.0\linewidth]{figures/so100_main.jpg}
%     \caption{\small{\textbf{Evaluation results on SO100 platform.} We evaluate EmbodiedMAE and SOTA VFM baselines under an RGB-only setting. Each task is tested over 10 trials.}}
%     \label{fig:so100_score}
% \end{figure}

% \begin{figure}[t]
%     \centering
%     \includegraphics[width=1.0\linewidth]{figures/score_xarm.png}
%     \caption{\small{\textbf{Evaluation results on xArm platform.} We evaluate EmbodiedMAE and SOTA VFM baselines under RGB-only, RGBD, and PC settings. Each task is tested over 10 trials.}}
%     \label{fig:xarm_score}
%     \vspace{-10pt}
% \end{figure}

\begin{figure}[t]
    \centering
    \includegraphics[width=1.0\linewidth]{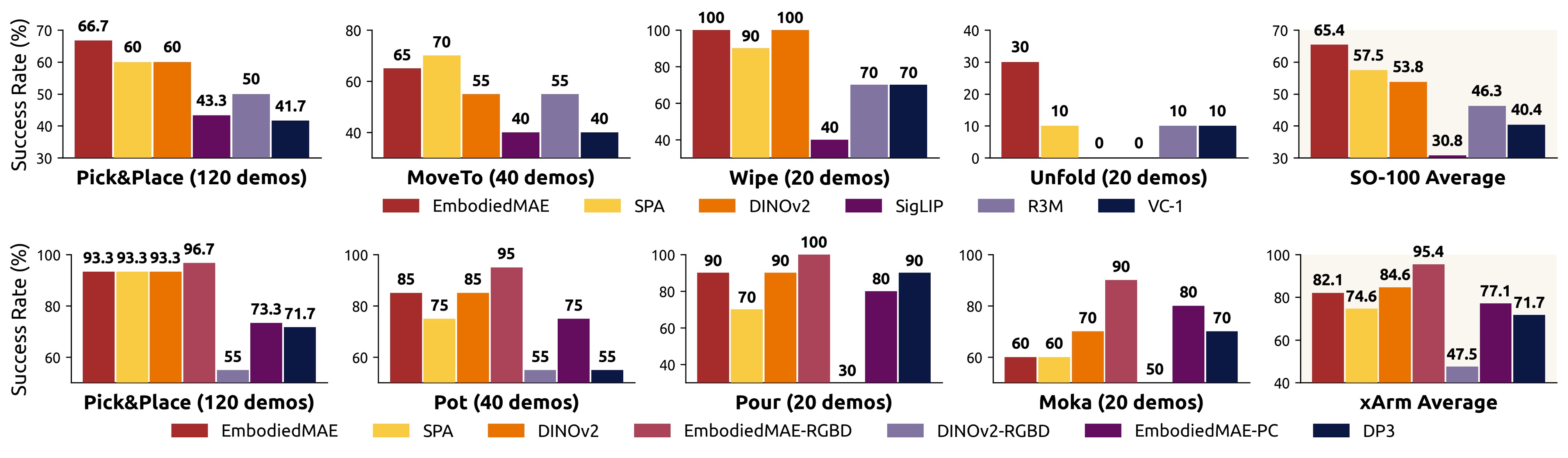}
    \vspace{-15pt}
    \caption{\small{\textbf{Evaluation results on SO100 and xArm platforms.} Each task is evaluated across 10 trials. On the SO100 platform, our model outperformed all baselines in the RGB-only setting. On the xArm platform, our model achieved comparable performance to SOTA baselines in the RGB-only setting, while significantly surpassing baselines in both RGBD and Point Cloud settings.}}
    \label{fig:xarm_so100_score}
    \vspace{-10pt}
\end{figure}

To further assess generalization in practical settings, we conduct real-world evaluations on two robot platforms: the low-cost, open-source SO100 \citep{cadene2024lerobot} and the high-performance xArm. We show quantitative results in \Cref{fig:xarm_so100_score}, and rollout visualizations in \Cref{fig:real_robot_demos}.

\textcolor{Maroon}{\textbf{\textit{Finding 1:}}} \textbf{EmbodiedMAE maintains SOTA performance in real-world robot manipulation.} EmbodiedMAE consistently achieves SOTA performance across real-world manipulation tasks, particularly those requiring strong spatial understanding. With multi-modal inputs, EmbodiedMAE further improves policy learning performance: EmbodiedMAE-RGBD and EmbodiedMAE-PC both surpass naïve fusion baselines such as DINOv2-RGBD (\Cref{appendix:dino_rgbd}) and DP3 \citep{Ze2024DP3}, highlighting the effectiveness of our design in promoting robust 3D perception for real-world robotics applications.

\textcolor{Maroon}{\textbf{\textit{Finding 2:}}} \textbf{3D information plays a critical role in robot manipulation.} Incorporating 3D inputs significantly improves task success rates. We observe that most failures in baseline models stem from localization errors, causing object loss, grasp failures, or collisions. EmbodiedMAE-RGBD, benefiting from enhanced spatial understanding, avoids these issues more reliably (see \Cref{fig:real_robot_demos}). The choice of 3D modality also matters. Although prior works \citep{li2025pointvla, Ze2024DP3, zhu2024pointcloudmatters} have highlighted the compactness and training efficiency of point cloud (PC) representations, we find their practical effectiveness is hindered by sensor noise from object reflectivity and lighting variations. Consequently, in our experiments, PC-based policies even underperform RGB-only inputs. In contrast, the RGBD setting, where depth serves as an auxiliary cue, yields better performance and is more robust to depth noise. This suggests that effective post-processing of point clouds is essential for leveraging them reliably; otherwise, RGBD inputs offer a more dependable alternative.

\subsection{Ablation Studies}

Due to the prohibitive cost of ViT-Giant pre-training, our ablation studies focus on model distillation insights. We evaluate masking ratio, feature alignment, and loss ratio on the LIBERO benchmark, reporting average success rates in \Cref{tab:ablation}, with default settings underlined. \textbf{(1) Masking Ratio}: Our default configuration sets 60 unmasked patches, approximately masking ratio of 90\%. We test 70\%, 80\%, and 100\% ratios (100\% representing training with only feature alignment loss). Results indicate performance insensitivity to masking ratio, though ratios <100\% perform better, suggesting feature alignment's predominant role while mask autoencoding provides additional benefits. \textbf{(2) Feature Alignment}: By default, we implement feature alignment at three positions (see \Cref{sec:distill}). Sequential removal of alignment points reveals diminishing impact from Top to Bottom, with each component contributing positively to model performance. \textbf{(3) Loss Ratio}: With default $\beta=1$, we test $\beta=0.5/2.0/4.0$. Results show performance robustness across $\beta$ values, with slight degradation at $\beta<1.0$, confirming feature alignment necessity, consistent with findings in \citep{bai2022dmae}.

\begin{wraptable}{r}{0.4\textwidth}
\centering
\small
\setlength{\tabcolsep}{3pt}
\scalebox{0.68}{
\begin{tabular}{l|cccc}
\toprule
Masking Ratio     & 0.7        & 0.8        & {\ul 0.9}     & 1.0  \\
\midrule
                  & 92.2       & 91.2       & 92.4    & 90.1 \\
\midrule
Feature Alignment & w/o Bottom & w/o Middle & w/o Top & {\ul All}  \\
\midrule
                  & 91.4       & 88.5       & 74.4    & 92.4 \\
\midrule
Loss Ratio $\beta$        & 0.5        & {\ul 1}          & 2       & 4    \\
\midrule
                  & 90.8       & 92.4       & 91.1    & 92.2 \\
\bottomrule
\end{tabular}}
\caption{\small{\textbf{Ablation study on LIBERO.} We conduct ablation experiments on masking ratio, feature alignment, and loss ratio on the LIBERO benchmark and report the average success rate.}}
\label{tab:ablation}
\end{wraptable}

\section{Related Works}

\textbf{Vision Foundation Models} are models trained on large-scale data in a self-supervised or semi-supervised manner that can be adapted for several other downstream tasks~\citep{bommasani2022opportunitiesrisksfoundationmodels}. Beyond conventional image classification, these models have shown strong transfer capabilities to tasks such as depth estimation \citep{depth_anything_v1, depth_anything_v2, Weinzaepfel2023crocov2}, semantic segmentation, and robot control \citep{octo_2023, kim2024openvla, liu2025rdt, kim2025openvlaoft}. Common pre-training techniques include contrastive learning \citep{he2019moco, chen2020mocov2, chen2021mocov3}, masked autoencoding \citep{bai2022dmae, tong2022videomae2, wang2023videomaev2, feichtenhofer2022videomae, he2022masked}, self-distillation \citep{oquab2024dinov, caron2021emerging}, and CLIP-style language-image contrastive learning \citep{zhai2023sigmoid, radford2021clip}. VFMs greatly improve AI systems' visual understanding.

\textbf{Visual Representations for Embodied AI} are crucial for enabling agents to perceive and interact with the physical world. Embodied perception must model robot-object interactions in dynamic environments, which general-purpose VFMs trained on static images often lack. Several recent methods have attempted to bridge this gap by training models directly on robot datasets. However, the limited scale and quality of embodied data hinder their generalization. These embodied-specific models often fail to generalize as well as VFMs trained on diverse in-the-wild datasets. As a result, many VLA models still rely on general-purpose VFMs like DINOv2 \citep{oquab2024dinov, kim2024openvla, kim2025openvlaoft} and SigLIP \citep{zhai2023sigmoid, liu2025rdt, kim2024openvla} for better generalization, prompting the need for dedicated large-scale embodied VFM pretraining.

\textbf{3D Robot Learning} has proven effective in improving both embodied agents' training efficiency and manipulation success rate \citep{Ze2024DP3, li2025pointvla, zhu2024pointcloudmatters}. Properly introducing 3D visual inputs, such as depth or point clouds, often leads to better spatial understanding compared to RGB-only inputs. However, naively incorporating 3D information, e.g., adding an extra depth channel, may severely degenerate the model's performance. Scalable native 3D multi-modal models remain largely absent in the current research landscape. EmbodiedMAE aims to address this gap by pre-training VFMs on large-scale, embodied-specific datasets to facilitate the development of scalable and effective 3D VLA models.

\section{Conclusion, Limitations, and Future Works}

In this work, we introduce EmbodiedMAE, a unified 3D multi-modal representation learning framework designed for robot learning. We first construct DROID-3D, a high-quality, large-scale 3D robot manipulation DROID supplement dataset, containing 76K trajectories. Then we propose a multi-modal masked autoencoder architecture that fuses RGB, depth, and point cloud inputs through stochastic masking and cross-modal decoding. Trained on DROID-3D, our model, EmbodiedMAE, demonstrates superior spatial understanding, strong multi-modal fusion ability, and effective scaling behavior. It outperforms strong VFM baselines across 70 simulation tasks and 20 real-world tasks on two robot platforms (SO100 and xArm). We believe both the DROID-3D dataset and EmbodiedMAE provide a valuable resource for 3D robot learning research. Despite the strong performance, EmbodiedMAE remains solely a vision backbone and does not natively support language instruction as input. A promising future direction is to fully leverage the language and action annotations available in the DROID-3D dataset to train a vision-language backbone, or even develop a multi-modal VLA model for instruction-following general embodied agents.

\bibliography{neurips_2025}

\begin{thebibliography}{49}
\providecommand{\natexlab}[1]{#1}
\providecommand{\url}[1]{\texttt{#1}}
\expandafter\ifx\csname urlstyle\endcsname\relax
  \providecommand{\doi}[1]{doi: #1}\else
  \providecommand{\doi}{doi: \begingroup \urlstyle{rm}\Url}\fi

\bibitem[Bachmann et~al.(2022)Bachmann, Mizrahi, Atanov, and Zamir]{bachmann2022multimae}
Roman Bachmann, David Mizrahi, Andrei Atanov, and Amir Zamir.
\newblock {MultiMAE}: Multi-modal multi-task masked autoencoders.
\newblock In \emph{European Conference on Computer Vision, {ECCV}}, 2022.

\bibitem[Bai et~al.(2023)Bai, Wang, Xiao, Wei, Wang, Yuille, Zhou, and Xie]{bai2022dmae}
Yutong Bai, Zeyu Wang, Junfei Xiao, Chen Wei, Huiyu Wang, Alan Yuille, Yuyin Zhou, and Cihang Xie.
\newblock Masked autoencoders enable efficient knowledge distillers.
\newblock In \emph{Conference on Computer Vision and Pattern Recognition, {CVPR}}, 2023.

\bibitem[Black et~al.(2024)Black, Brown, Driess, Esmail, Equi, Finn, Fusai, Groom, Hausman, Ichter, Jakubczak, Jones, Ke, Levine, Li-Bell, Mothukuri, Nair, Pertsch, Shi, Tanner, Vuong, Walling, Wang, and Zhilinsky]{black2024pi0}
Kevin Black, Noah Brown, Danny Driess, Adnan Esmail, Michael Equi, Chelsea Finn, Niccolo Fusai, Lachy Groom, Karol Hausman, Brian Ichter, Szymon Jakubczak, Tim Jones, Liyiming Ke, Sergey Levine, Adrian Li-Bell, Mohith Mothukuri, Suraj Nair, Karl Pertsch, Lucy~Xiaoyang Shi, James Tanner, Quan Vuong, Anna Walling, Haohuan Wang, and Ury Zhilinsky.
\newblock $\pi_0$: A vision-language-action flow model for general robot control.
\newblock \emph{arXiv preprint arXiv:2410.24164}, 2024.

\bibitem[Bommasani et~al.(2022)Bommasani, Hudson, Adeli, Altman, Arora, von Arx, Bernstein, Bohg, Bosselut, Brunskill, Brynjolfsson, Buch, Card, Castellon, Chatterji, Chen, Creel, Davis, Demszky, Donahue, Doumbouya, Durmus, Ermon, Etchemendy, Ethayarajh, Fei-Fei, Finn, Gale, Gillespie, Goel, Goodman, Grossman, Guha, Hashimoto, Henderson, Hewitt, Ho, Hong, Hsu, Huang, Icard, Jain, Jurafsky, Kalluri, Karamcheti, Keeling, Khani, Khattab, Koh, Krass, Krishna, Kuditipudi, Kumar, Ladhak, Lee, Lee, Leskovec, Levent, Li, Li, Ma, Malik, Manning, Mirchandani, Mitchell, Munyikwa, Nair, Narayan, Narayanan, Newman, Nie, Niebles, Nilforoshan, Nyarko, Ogut, Orr, Papadimitriou, Park, Piech, Portelance, Potts, Raghunathan, Reich, Ren, Rong, Roohani, Ruiz, Ryan, Ré, Sadigh, Sagawa, Santhanam, Shih, Srinivasan, Tamkin, Taori, Thomas, Tramèr, Wang, Wang, Wu, Wu, Wu, Xie, Yasunaga, You, Zaharia, Zhang, Zhang, Zhang, Zhang, Zheng, Zhou, and Liang]{bommasani2022opportunitiesrisksfoundationmodels}
Rishi Bommasani, Drew~A. Hudson, Ehsan Adeli, Russ Altman, Simran Arora, Sydney von Arx, Michael~S. Bernstein, Jeannette Bohg, Antoine Bosselut, Emma Brunskill, Erik Brynjolfsson, Shyamal Buch, Dallas Card, Rodrigo Castellon, Niladri Chatterji, Annie Chen, Kathleen Creel, Jared~Quincy Davis, Dora Demszky, Chris Donahue, Moussa Doumbouya, Esin Durmus, Stefano Ermon, John Etchemendy, Kawin Ethayarajh, Li~Fei-Fei, Chelsea Finn, Trevor Gale, Lauren Gillespie, Karan Goel, Noah Goodman, Shelby Grossman, Neel Guha, Tatsunori Hashimoto, Peter Henderson, John Hewitt, Daniel~E. Ho, Jenny Hong, Kyle Hsu, Jing Huang, Thomas Icard, Saahil Jain, Dan Jurafsky, Pratyusha Kalluri, Siddharth Karamcheti, Geoff Keeling, Fereshte Khani, Omar Khattab, Pang~Wei Koh, Mark Krass, Ranjay Krishna, Rohith Kuditipudi, Ananya Kumar, Faisal Ladhak, Mina Lee, Tony Lee, Jure Leskovec, Isabelle Levent, Xiang~Lisa Li, Xuechen Li, Tengyu Ma, Ali Malik, Christopher~D. Manning, Suvir Mirchandani, Eric Mitchell, Zanele Munyikwa, Suraj Nair,
  Avanika Narayan, Deepak Narayanan, Ben Newman, Allen Nie, Juan~Carlos Niebles, Hamed Nilforoshan, Julian Nyarko, Giray Ogut, Laurel Orr, Isabel Papadimitriou, Joon~Sung Park, Chris Piech, Eva Portelance, Christopher Potts, Aditi Raghunathan, Rob Reich, Hongyu Ren, Frieda Rong, Yusuf Roohani, Camilo Ruiz, Jack Ryan, Christopher Ré, Dorsa Sadigh, Shiori Sagawa, Keshav Santhanam, Andy Shih, Krishnan Srinivasan, Alex Tamkin, Rohan Taori, Armin~W. Thomas, Florian Tramèr, Rose~E. Wang, William Wang, Bohan Wu, Jiajun Wu, Yuhuai Wu, Sang~Michael Xie, Michihiro Yasunaga, Jiaxuan You, Matei Zaharia, Michael Zhang, Tianyi Zhang, Xikun Zhang, Yuhui Zhang, Lucia Zheng, Kaitlyn Zhou, and Percy Liang.
\newblock On the opportunities and risks of foundation models.
\newblock In \emph{arXiv preprint arXiv:2108.07258}, 2022.

\bibitem[Cadene et~al.(2024)Cadene, Alibert, Soare, Gallouedec, Zouitine, and Wolf]{cadene2024lerobot}
Remi Cadene, Simon Alibert, Alexander Soare, Quentin Gallouedec, Adil Zouitine, and Thomas Wolf.
\newblock Lerobot: State-of-the-art machine learning for real-world robotics in pytorch.
\newblock \url{https://github.com/huggingface/lerobot}, 2024.

\bibitem[Caron et~al.(2021)Caron, Touvron, Misra, J\'egou, Mairal, Bojanowski, and Joulin]{caron2021emerging}
Mathilde Caron, Hugo Touvron, Ishan Misra, Herv\'e J\'egou, Julien Mairal, Piotr Bojanowski, and Armand Joulin.
\newblock Emerging properties in self-supervised vision transformers.
\newblock In \emph{Proceedings of the International Conference on Computer Vision, {ICCV}}, 2021.

\bibitem[Chen et~al.(2020)Chen, Fan, Girshick, and He]{chen2020mocov2}
Xinlei Chen, Haoqi Fan, Ross Girshick, and Kaiming He.
\newblock Improved baselines with momentum contrastive learning.
\newblock In \emph{arXiv preprint arXiv:2003.04297}, 2020.

\bibitem[Chen* et~al.(2021)Chen*, Xie*, and He]{chen2021mocov3}
Xinlei Chen*, Saining Xie*, and Kaiming He.
\newblock An empirical study of training self-supervised vision transformers.
\newblock In \emph{arXiv preprint arXiv:2104.02057}, 2021.

\bibitem[Chi et~al.(2023)Chi, Feng, Du, Xu, Cousineau, Burchfiel, and Song]{chi2023diffusionpolicy}
Cheng Chi, Siyuan Feng, Yilun Du, Zhenjia Xu, Eric Cousineau, Benjamin Burchfiel, and Shuran Song.
\newblock Diffusion policy: Visuomotor policy learning via action diffusion.
\newblock In \emph{Proceedings of Robotics: Science and Systems, {RSS}}, 2023.

\bibitem[Dong et~al.(2024)Dong, Yuan, HAO, Ni, Ma, Li, and ZHENG]{dong2024cleandiffuser}
Zibin Dong, Yifu Yuan, Jianye HAO, Fei Ni, Yi~Ma, Pengyi Li, and YAN ZHENG.
\newblock Cleandiffuser: An easy-to-use modularized library for diffusion models in decision making.
\newblock In \emph{The Thirty-eight Conference on Neural Information Processing Systems Datasets and Benchmarks Track, {NIPS}}, 2024.

\bibitem[Dosovitskiy et~al.(2021)Dosovitskiy, Beyer, Kolesnikov, Weissenborn, Zhai, Unterthiner, Dehghani, Minderer, Heigold, Gelly, Uszkoreit, and Houlsby]{dosovitskiy2021vit}
Alexey Dosovitskiy, Lucas Beyer, Alexander Kolesnikov, Dirk Weissenborn, Xiaohua Zhai, Thomas Unterthiner, Mostafa Dehghani, Matthias Minderer, Georg Heigold, Sylvain Gelly, Jakob Uszkoreit, and Neil Houlsby.
\newblock An image is worth 16x16 words: Transformers for image recognition at scale.
\newblock In \emph{International Conference on Learning Representations, {ICLR}}, 2021.

\bibitem[Fang et~al.(2023)Fang, Fang, Tang, Liu, Wang, Zhu, and Lu]{fang2023rh20t}
Hao-Shu Fang, Hongjie Fang, Zhenyu Tang, Jirong Liu, Junbo Wang, Haoyi Zhu, and Cewu Lu.
\newblock Rh20t: A robotic dataset for learning diverse skills in one-shot.
\newblock In \emph{RSS 2023 Workshop on Learning for Task and Motion Planning, {RSS}}, 2023.

\bibitem[Feichtenhofer et~al.(2022)Feichtenhofer, Fan, Li, and He]{feichtenhofer2022videomae}
Christoph Feichtenhofer, Haoqi Fan, Yanghao Li, and Kaiming He.
\newblock Masked autoencoders as spatiotemporal learners.
\newblock In \emph{Advances in Neural Information Processing Systems, {NIPS}}, 2022.

\bibitem[He et~al.(2019)He, Fan, Wu, Xie, and Girshick]{he2019moco}
Kaiming He, Haoqi Fan, Yuxin Wu, Saining Xie, and Ross Girshick.
\newblock Momentum contrast for unsupervised visual representation learning.
\newblock In \emph{arXiv preprint arXiv:1911.05722}, 2019.

\bibitem[He et~al.(2022)He, Chen, Xie, Li, Doll{\'{a}}r, and Girshick]{he2022masked}
Kaiming He, Xinlei Chen, Saining Xie, Yanghao Li, Piotr Doll{\'{a}}r, and Ross~B. Girshick.
\newblock Masked autoencoders are scalable vision learners.
\newblock In \emph{{IEEE/CVF} Conference on Computer Vision and Pattern Recognition, {CVPR}}, 2022.

\bibitem[Huang et~al.(2023)Huang, Peng, He, Yang, Zhou, and Ouyang]{huang2023ponder}
Di~Huang, Sida Peng, Tong He, Honghui Yang, Xiaowei Zhou, and Wanli Ouyang.
\newblock Ponder: Point cloud pre-training via neural rendering.
\newblock In \emph{Proceedings of the IEEE/CVF International Conference on Computer Vision, {ICCV}}, 2023.

\bibitem[Ke et~al.(2024)Ke, Gkanatsios, and Fragkiadaki]{ke20243ddiffusionactor}
Tsung-Wei Ke, Nikolaos Gkanatsios, and Katerina Fragkiadaki.
\newblock 3d diffuser actor: Policy diffusion with 3d scene representations.
\newblock In \emph{8th Annual Conference on Robot Learning, {CoRL}}, 2024.

\bibitem[Khazatsky et~al.(2024)Khazatsky, Pertsch, Nair, Balakrishna, Dasari, Karamcheti, Nasiriany, Srirama, Chen, Ellis, Fagan, Hejna, Itkina, Lepert, Ma, Miller, Wu, Belkhale, Dass, Ha, Jain, Lee, Lee, Memmel, Park, Radosavovic, Wang, Zhan, Black, Chi, Hatch, Lin, Lu, Mercat, Rehman, Sanketi, Sharma, Simpson, Vuong, Walke, Wulfe, Xiao, Yang, Yavary, Zhao, Agia, Baijal, Castro, Chen, Chen, Chung, Drake, Foster, Gao, Herrera, Heo, Hsu, Hu, Jackson, Le, Li, Lin, Ma, Maddukuri, Mirchandani, Morton, Nguyen, O'Neill, Scalise, Seale, Son, Tian, Tran, Wang, Wu, Xie, Yang, Yin, Zhang, Bastani, Berseth, Bohg, Goldberg, Gupta, Gupta, Jayaraman, Lim, Malik, Mart{\'\i}n-Mart{\'\i}n, Ramamoorthy, Sadigh, Song, Wu, Yip, Zhu, Kollar, Levine, and Finn]{khazatsky2024droid}
Alexander Khazatsky, Karl Pertsch, Suraj Nair, Ashwin Balakrishna, Sudeep Dasari, Siddharth Karamcheti, Soroush Nasiriany, Mohan~Kumar Srirama, Lawrence~Yunliang Chen, Kirsty Ellis, Peter~David Fagan, Joey Hejna, Masha Itkina, Marion Lepert, Yecheng~Jason Ma, Patrick~Tree Miller, Jimmy Wu, Suneel Belkhale, Shivin Dass, Huy Ha, Arhan Jain, Abraham Lee, Youngwoon Lee, Marius Memmel, Sungjae Park, Ilija Radosavovic, Kaiyuan Wang, Albert Zhan, Kevin Black, Cheng Chi, Kyle~Beltran Hatch, Shan Lin, Jingpei Lu, Jean Mercat, Abdul Rehman, Pannag~R Sanketi, Archit Sharma, Cody Simpson, Quan Vuong, Homer~Rich Walke, Blake Wulfe, Ted Xiao, Jonathan~Heewon Yang, Arefeh Yavary, Tony~Z. Zhao, Christopher Agia, Rohan Baijal, Mateo~Guaman Castro, Daphne Chen, Qiuyu Chen, Trinity Chung, Jaimyn Drake, Ethan~Paul Foster, Jensen Gao, David~Antonio Herrera, Minho Heo, Kyle Hsu, Jiaheng Hu, Donovon Jackson, Charlotte Le, Yunshuang Li, Roy Lin, Zehan Ma, Abhiram Maddukuri, Suvir Mirchandani, Daniel Morton, Tony~Khuong Nguyen,
  Abigail O'Neill, Rosario Scalise, Derick Seale, Victor Son, Stephen Tian, Emi Tran, Andrew~E. Wang, Yilin Wu, Annie Xie, Jingyun Yang, Patrick Yin, Yunchu Zhang, Osbert Bastani, Glen Berseth, Jeannette Bohg, Ken Goldberg, Abhinav Gupta, Abhishek Gupta, Dinesh Jayaraman, Joseph~J Lim, Jitendra Malik, Roberto Mart{\'\i}n-Mart{\'\i}n, Subramanian Ramamoorthy, Dorsa Sadigh, Shuran Song, Jiajun Wu, Michael~C. Yip, Yuke Zhu, Thomas Kollar, Sergey Levine, and Chelsea Finn.
\newblock {DROID}: A large-scale in-the-wild robot manipulation dataset.
\newblock In \emph{RSS 2024 Workshop: Data Generation for Robotics, {RSS}}, 2024.

\bibitem[Kim et~al.(2024)Kim, Pertsch, Karamcheti, Xiao, Balakrishna, Nair, Rafailov, Foster, Sanketi, Vuong, Kollar, Burchfiel, Tedrake, Sadigh, Levine, Liang, and Finn]{kim2024openvla}
Moo~Jin Kim, Karl Pertsch, Siddharth Karamcheti, Ted Xiao, Ashwin Balakrishna, Suraj Nair, Rafael Rafailov, Ethan~P Foster, Pannag~R Sanketi, Quan Vuong, Thomas Kollar, Benjamin Burchfiel, Russ Tedrake, Dorsa Sadigh, Sergey Levine, Percy Liang, and Chelsea Finn.
\newblock Open{VLA}: An open-source vision-language-action model.
\newblock In \emph{8th Annual Conference on Robot Learning, {CoRL}}, 2024.

\bibitem[Kim et~al.(2025)Kim, Finn, and Liang]{kim2025openvlaoft}
Moo~Jin Kim, Chelsea Finn, and Percy Liang.
\newblock Fine-tuning vision-language-action models: Optimizing speed and success.
\newblock In \emph{arXiv preprint arXiv:2502.19645}, 2025.

\bibitem[Li et~al.(2025{\natexlab{a}})Li, Wen, Peng, Peng, Feng, and Zhu]{li2025pointvla}
Chengmeng Li, Junjie Wen, Yan Peng, Yaxin Peng, Feifei Feng, and Yichen Zhu.
\newblock Pointvla: Injecting the 3d world into vision-language-action models.
\newblock In \emph{arXiv preprint arXiv:2503.07511}, 2025{\natexlab{a}}.

\bibitem[Li et~al.(2025{\natexlab{b}})Li, Shao, Zhang, Wang, Brunswic, Zhou, Dong, Guo, Li, Chen, Wang, and Hao]{li2025survey}
Yinchuan Li, Xinyu Shao, Jianping Zhang, Haozhi Wang, Leo~Maxime Brunswic, Kaiwen Zhou, Jiqian Dong, Kaiyang Guo, Xiu Li, Zhitang Chen, Jun Wang, and Jianye Hao.
\newblock Generative models in decision making: A survey.
\newblock In \emph{arXiv preprint arXiv:2502.17100}, 2025{\natexlab{b}}.

\bibitem[Liu et~al.(2023)Liu, Zhu, Gao, Feng, qiang liu, Zhu, and Stone]{liu2023libero}
Bo~Liu, Yifeng Zhu, Chongkai Gao, Yihao Feng, qiang liu, Yuke Zhu, and Peter Stone.
\newblock {LIBERO}: Benchmarking knowledge transfer for lifelong robot learning.
\newblock In \emph{Thirty-seventh Conference on Neural Information Processing Systems Datasets and Benchmarks Track, {NIPS}}, 2023.

\bibitem[Liu et~al.(2025)Liu, Wu, Li, Tan, Chen, Wang, Xu, Su, and Zhu]{liu2025rdt}
Songming Liu, Lingxuan Wu, Bangguo Li, Hengkai Tan, Huayu Chen, Zhengyi Wang, Ke~Xu, Hang Su, and Jun Zhu.
\newblock {RDT}-1b: a diffusion foundation model for bimanual manipulation.
\newblock In \emph{The Thirteenth International Conference on Learning Representations, {ICLR}}, 2025.

\bibitem[Majumdar et~al.(2023)Majumdar, Yadav, Arnaud, Ma, Chen, Silwal, Jain, Berges, Wu, Vakil, Abbeel, Malik, Batra, Lin, Maksymets, Rajeswaran, and Meier]{majumdar2023vc1}
Arjun Majumdar, Karmesh Yadav, Sergio Arnaud, Yecheng~Jason Ma, Claire Chen, Sneha Silwal, Aryan Jain, Vincent-Pierre Berges, Tingfan Wu, Jay Vakil, Pieter Abbeel, Jitendra Malik, Dhruv Batra, Yixin Lin, Oleksandr Maksymets, Aravind Rajeswaran, and Franziska Meier.
\newblock Where are we in the search for an artificial visual cortex for embodied intelligence?
\newblock In \emph{Thirty-seventh Conference on Neural Information Processing Systems, {NIPS}}, 2023.

\bibitem[Nair et~al.(2022)Nair, Rajeswaran, Kumar, Finn, and Gupta]{nair2022r3m}
Suraj Nair, Aravind Rajeswaran, Vikash Kumar, Chelsea Finn, and Abhinav Gupta.
\newblock R3m: A universal visual representation for robot manipulation.
\newblock In \emph{6th Annual Conference on Robot Learning, {CoRL}}, 2022.

\bibitem[{Octo Model Team} et~al.(2024){Octo Model Team}, Ghosh, Walke, Pertsch, Black, Mees, Dasari, Hejna, Xu, Luo, Kreiman, Tan, Sanketi, Vuong, Xiao, Sadigh, Finn, and Levine]{octo_2023}
{Octo Model Team}, Dibya Ghosh, Homer Walke, Karl Pertsch, Kevin Black, Oier Mees, Sudeep Dasari, Joey Hejna, Charles Xu, Jianlan Luo, Tobias Kreiman, {You Liang} Tan, Pannag Sanketi, Quan Vuong, Ted Xiao, Dorsa Sadigh, Chelsea Finn, and Sergey Levine.
\newblock Octo: An open-source generalist robot policy.
\newblock In \emph{Proceedings of Robotics: Science and Systems, {RSS}}, 2024.

\bibitem[Oquab et~al.(2024)Oquab, Darcet, Moutakanni, Vo, Szafraniec, Khalidov, Fernandez, HAZIZA, Massa, El-Nouby, Assran, Ballas, Galuba, Howes, Huang, Li, Misra, Rabbat, Sharma, Synnaeve, Xu, Jegou, Mairal, Labatut, Joulin, and Bojanowski]{oquab2024dinov}
Maxime Oquab, Timoth{\'e}e Darcet, Th{\'e}o Moutakanni, Huy~V. Vo, Marc Szafraniec, Vasil Khalidov, Pierre Fernandez, Daniel HAZIZA, Francisco Massa, Alaaeldin El-Nouby, Mido Assran, Nicolas Ballas, Wojciech Galuba, Russell Howes, Po-Yao Huang, Shang-Wen Li, Ishan Misra, Michael Rabbat, Vasu Sharma, Gabriel Synnaeve, Hu~Xu, Herve Jegou, Julien Mairal, Patrick Labatut, Armand Joulin, and Piotr Bojanowski.
\newblock {DINO}v2: Learning robust visual features without supervision.
\newblock \emph{Transactions on Machine Learning Research, {TMLR}}, 2024.

\bibitem[Pang et~al.(2022)Pang, Wang, Tay, Liu, Tian, and Yuan]{pang2022pointmae}
Yatian Pang, Wenxiao Wang, Francis~EH Tay, Wei Liu, Yonghong Tian, and Li~Yuan.
\newblock Masked autoencoders for point cloud self-supervised learning.
\newblock In \emph{European Conference on Computer Vision, {ECCV}}, 2022.

\bibitem[Peebles and Xie(2022)]{Peebles2022DiT}
William Peebles and Saining Xie.
\newblock Scalable diffusion models with transformers.
\newblock In \emph{arXiv preprint arXiv:2212.09748}, 2022.

\bibitem[Qian et~al.(2022)Qian, Li, Peng, Mai, Hammoud, Elhoseiny, and Ghanem]{qian2022pointnext}
Guocheng Qian, Yuchen Li, Houwen Peng, Jinjie Mai, Hasan Abed Al~Kader Hammoud, Mohamed Elhoseiny, and Bernard Ghanem.
\newblock Pointnext: Revisiting pointnet++ with improved training and scaling strategies.
\newblock In \emph{Advances in Neural Information Processing Systems, {NIPS}}, 2022.

\bibitem[Qu et~al.(2025)Qu, Song, Chen, Yao, Ye, Ding, Wang, Gu, Zhao, Wang, and Li]{qu2025spatialvla}
Delin Qu, Haoming Song, Qizhi Chen, Yuanqi Yao, Xinyi Ye, Yan Ding, Zhigang Wang, JiaYuan Gu, Bin Zhao, Dong Wang, and Xuelong Li.
\newblock Spatialvla: Exploring spatial representations for visual-language-action model.
\newblock In \emph{arXiv preprint arXiv:2501.15830}, 2025.

\bibitem[Radford et~al.(2021)Radford, Kim, Hallacy, Ramesh, Goh, Agarwal, Sastry, Askell, Mishkin, Clark, Krueger, and Sutskever]{radford2021clip}
Alec Radford, Jong~Wook Kim, Chris Hallacy, Aditya Ramesh, Gabriel Goh, Sandhini Agarwal, Girish Sastry, Amanda Askell, Pamela Mishkin, Jack Clark, Gretchen Krueger, and Ilya Sutskever.
\newblock Learning transferable visual models from natural language supervision.
\newblock In \emph{arXiv preprint arXiv:2103.00020}, 2021.

\bibitem[Tong et~al.(2022)Tong, Song, Wang, and Wang]{tong2022videomae2}
Zhan Tong, Yibing Song, Jue Wang, and Limin Wang.
\newblock Video{MAE}: Masked autoencoders are data-efficient learners for self-supervised video pre-training.
\newblock In \emph{Advances in Neural Information Processing Systems, {NIPS}}, 2022.

\bibitem[Touvron et~al.(2021)Touvron, Cord, Douze, Massa, Sablayrolles, and Jegou]{hugo2021deit}
Hugo Touvron, Matthieu Cord, Matthijs Douze, Francisco Massa, Alexandre Sablayrolles, and Herve Jegou.
\newblock Training data-efficient image transformers \& distillation through attention.
\newblock In \emph{Proceedings of the 38th International Conference on Machine Learning, {ICML}}, 2021.

\bibitem[Vuong et~al.(2023)Vuong, Levine, Walke, Pertsch, Singh, Doshi, Xu, Luo, Tan, Shah, Finn, Du, Kim, Khazatsky, Yang, Zhao, Goldberg, Hoque, Chen, Adebola, Sukhatme, Salhotra, Dass, Pinto, Cui, Haldar, Rai, Shafiullah, Zhu, Zhu, Nasiriany, Song, Chi, Pan, Burgard, Mees, Huang, Pathak, Bahl, Mendonca, Zhou, Srirama, Dasari, Lu, Fang, Fang, Christensen, Tomizuka, Zhan, Ding, Xu, Zhu, Tian, Lee, Sadigh, Cui, Belkhale, Sundaresan, Darrell, Malik, Radosavovic, Bohg, Srinivasan, Wang, Hansen, Wu, Yan, Su, Gu, Li, Suenderhauf, Rana, Burgess-Limerick, Ceola, Kawaharazuka, Kanazawa, Matsushima, Matsuo, Iwasawa, Furuta, Oh, Harada, Osa, Tang, Kroemer, Sharma, Zhang, Kim, Cho, Han, Kim, Lim, Johns, Palo, Stulp, Raffin, Bustamante, Silv{\'e}rio, Padalkar, Peters, Sch{\"o}lkopf, B{\"u}chler, Schneider, Guist, Wu, Tian, Shi, Li, Wang, Zhang, Amor, Zhou, Majd, Ott, Schiavi, Mart{\'\i}n-Mart{\'\i}n, Shah, Bisk, Bingham, Yu, Jain, Xiao, Hausman, Chan, Herzog, Xu, Kirmani, Vanhoucke, Julian, Lee, Ding, Chebotar, Tan,
  Liang, Mordatch, Rao, Lu, Gopalakrishnan, Welker, Joshi, Devin, Irpan, Moore, Wahid, Wu, Chen, Wohlhart, Bewley, Zhou, Leal, Kalashnikov, Sanketi, Fu, Xu, Xu, brian ichter, Hsu, Xu, Brohan, Sermanet, Heess, Ahn, Rafailov, Pooley, Byrne, Davchev, Oslund, Schaal, Jain, Go, Xia, Tompson, Armstrong, and Driess]{vuong2023oxe}
Quan Vuong, Sergey Levine, Homer~Rich Walke, Karl Pertsch, Anikait Singh, Ria Doshi, Charles Xu, Jianlan Luo, Liam Tan, Dhruv Shah, Chelsea Finn, Max Du, Moo~Jin Kim, Alexander Khazatsky, Jonathan~Heewon Yang, Tony~Z. Zhao, Ken Goldberg, Ryan Hoque, Lawrence~Yunliang Chen, Simeon Adebola, Gaurav~S. Sukhatme, Gautam Salhotra, Shivin Dass, Lerrel Pinto, Zichen~Jeff Cui, Siddhant Haldar, Anant Rai, Nur Muhammad~Mahi Shafiullah, Yuke Zhu, Yifeng Zhu, Soroush Nasiriany, Shuran Song, Cheng Chi, Chuer Pan, Wolfram Burgard, Oier Mees, Chenguang Huang, Deepak Pathak, Shikhar Bahl, Russell Mendonca, Gaoyue Zhou, Mohan~Kumar Srirama, Sudeep Dasari, Cewu Lu, Hao-Shu Fang, Hongjie Fang, Henrik~I Christensen, Masayoshi Tomizuka, Wei Zhan, Mingyu Ding, Chenfeng Xu, Xinghao Zhu, Ran Tian, Youngwoon Lee, Dorsa Sadigh, Yuchen Cui, Suneel Belkhale, Priya Sundaresan, Trevor Darrell, Jitendra Malik, Ilija Radosavovic, Jeannette Bohg, Krishnan Srinivasan, Xiaolong Wang, Nicklas Hansen, Yueh-Hua Wu, Ge~Yan, Hao Su, Jiayuan Gu,
  Xuanlin Li, Niko Suenderhauf, Krishan Rana, Ben Burgess-Limerick, Federico Ceola, Kento Kawaharazuka, Naoaki Kanazawa, Tatsuya Matsushima, Yutaka Matsuo, Yusuke Iwasawa, Hiroki Furuta, Jihoon Oh, Tatsuya Harada, Takayuki Osa, Yujin Tang, Oliver Kroemer, Mohit Sharma, Kevin~Lee Zhang, Beomjoon Kim, Yoonyoung Cho, Junhyek Han, Jaehyung Kim, Joseph~J Lim, Edward Johns, Norman~Di Palo, Freek Stulp, Antonin Raffin, Samuel Bustamante, Jo{\~a}o Silv{\'e}rio, Abhishek Padalkar, Jan Peters, Bernhard Sch{\"o}lkopf, Dieter B{\"u}chler, Jan Schneider, Simon Guist, Jiajun Wu, Stephen Tian, Haochen Shi, Yunzhu Li, Yixuan Wang, Mingtong Zhang, Heni~Ben Amor, Yifan Zhou, Keyvan Majd, Lionel Ott, Giulio Schiavi, Roberto Mart{\'\i}n-Mart{\'\i}n, Rutav Shah, Yonatan Bisk, Jeffrey~T Bingham, Tianhe Yu, Vidhi Jain, Ted Xiao, Karol Hausman, Christine Chan, Alexander Herzog, Zhuo Xu, Sean Kirmani, Vincent Vanhoucke, Ryan Julian, Lisa Lee, Tianli Ding, Yevgen Chebotar, Jie Tan, Jacky Liang, Igor Mordatch, Kanishka Rao, Yao Lu,
  Keerthana Gopalakrishnan, Stefan Welker, Nikhil~J Joshi, Coline~Manon Devin, Alex Irpan, Sherry Moore, Ayzaan Wahid, Jialin Wu, Xi~Chen, Paul Wohlhart, Alex Bewley, Wenxuan Zhou, Isabel Leal, Dmitry Kalashnikov, Pannag~R Sanketi, Chuyuan Fu, Ying Xu, Sichun Xu, brian ichter, Jasmine Hsu, Peng Xu, Anthony Brohan, Pierre Sermanet, Nicolas Heess, Michael Ahn, Rafael Rafailov, Acorn Pooley, Kendra Byrne, Todor Davchev, Kenneth Oslund, Stefan Schaal, Ajinkya Jain, Keegan Go, Fei Xia, Jonathan Tompson, Travis Armstrong, and Danny Driess.
\newblock Open x-embodiment: Robotic learning datasets and {RT}-x models.
\newblock In \emph{Towards Generalist Robots: Learning Paradigms for Scalable Skill Acquisition, {CoRL}}, 2023.

\bibitem[Walke et~al.(2023)Walke, Black, Lee, Kim, Du, Zheng, Zhao, Hansen-Estruch, Vuong, He, Myers, Fang, Finn, and Levine]{walke2023bridgedata}
Homer Walke, Kevin Black, Abraham Lee, Moo~Jin Kim, Max Du, Chongyi Zheng, Tony Zhao, Philippe Hansen-Estruch, Quan Vuong, Andre He, Vivek Myers, Kuan Fang, Chelsea Finn, and Sergey Levine.
\newblock Bridgedata v2: A dataset for robot learning at scale.
\newblock In \emph{Conference on Robot Learning, {CoRL}}, 2023.

\bibitem[Wang et~al.(2023)Wang, Huang, Zhao, Tong, He, Wang, Wang, and Qiao]{wang2023videomaev2}
Limin Wang, Bingkun Huang, Zhiyu Zhao, Zhan Tong, Yinan He, Yi~Wang, Yali Wang, and Yu~Qiao.
\newblock Videomae v2: Scaling video masked autoencoders with dual masking.
\newblock In \emph{Conference on Computer Vision and Pattern Recognition, {CVPR}}, 2023.

\bibitem[Weinzaepfel et~al.(2023)Weinzaepfel, Lucas, Leroy, Cabon, Arora, Brégier, Csurka, Antsfeld, Chidlovskii, and Revaud]{Weinzaepfel2023crocov2}
Philippe Weinzaepfel, Thomas Lucas, Vincent Leroy, Yohann Cabon, Vaibhav Arora, Romain Brégier, Gabriela Csurka, Leonid Antsfeld, Boris Chidlovskii, and Jérôme Revaud.
\newblock Croco v2: Improved cross-view completion pre-training for stereo matching and optical flow.
\newblock In \emph{Proceedings of the IEEE/CVF International Conference on Computer Vision, {ICCV}}, 2023.

\bibitem[Wolf et~al.(2020)Wolf, Debut, Sanh, Chaumond, Delangue, Moi, Cistac, Rault, Louf, Funtowicz, Davison, Shleifer, von Platen, Ma, Jernite, Plu, Xu, Scao, Gugger, Drame, Lhoest, and Rush]{wolf-etal-2020-transformers}
Thomas Wolf, Lysandre Debut, Victor Sanh, Julien Chaumond, Clement Delangue, Anthony Moi, Pierric Cistac, Tim Rault, Rémi Louf, Morgan Funtowicz, Joe Davison, Sam Shleifer, Patrick von Platen, Clara Ma, Yacine Jernite, Julien Plu, Canwen Xu, Teven~Le Scao, Sylvain Gugger, Mariama Drame, Quentin Lhoest, and Alexander~M. Rush.
\newblock Transformers: State-of-the-art natural language processing.
\newblock In \emph{Proceedings of the 2020 Conference on Empirical Methods in Natural Language Processing: System Demonstrations, {EMNLP}}, 2020.

\bibitem[Yang et~al.(2024{\natexlab{a}})Yang, Kang, Huang, Xu, Feng, and Zhao]{depth_anything_v1}
Lihe Yang, Bingyi Kang, Zilong Huang, Xiaogang Xu, Jiashi Feng, and Hengshuang Zhao.
\newblock Depth anything: Unleashing the power of large-scale unlabeled data.
\newblock In \emph{Conference on Computer Vision and Pattern Recognition, {CVPR}}, 2024{\natexlab{a}}.

\bibitem[Yang et~al.(2024{\natexlab{b}})Yang, Kang, Huang, Zhao, Xu, Feng, and Zhao]{depth_anything_v2}
Lihe Yang, Bingyi Kang, Zilong Huang, Zhen Zhao, Xiaogang Xu, Jiashi Feng, and Hengshuang Zhao.
\newblock Depth anything v2.
\newblock In \emph{arXiv preprint arXiv:2406.09414}, 2024{\natexlab{b}}.

\bibitem[Yu et~al.(2019)Yu, Quillen, He, Julian, Hausman, Finn, and Levine]{yu2019metaworld}
Tianhe Yu, Deirdre Quillen, Zhanpeng He, Ryan Julian, Karol Hausman, Chelsea Finn, and Sergey Levine.
\newblock Meta-world: A benchmark and evaluation for multi-task and meta reinforcement learning.
\newblock In \emph{Conference on Robot Learning, {CoRL}}, 2019.

\bibitem[Ze et~al.(2024)Ze, Zhang, Zhang, Hu, Wang, and Xu]{Ze2024DP3}
Yanjie Ze, Gu~Zhang, Kangning Zhang, Chenyuan Hu, Muhan Wang, and Huazhe Xu.
\newblock 3d diffusion policy: Generalizable visuomotor policy learning via simple 3d representations.
\newblock In \emph{Proceedings of Robotics: Science and Systems, {RSS}}, 2024.

\bibitem[Zhai et~al.(2023)Zhai, Mustafa, Kolesnikov, and Beyer]{zhai2023sigmoid}
Xiaohua Zhai, Basil Mustafa, Alexander Kolesnikov, and Lucas Beyer.
\newblock Sigmoid loss for language image pre-training.
\newblock In \emph{{IEEE/CVF} International Conference on Computer Vision, {ICCV}}, 2023.

\bibitem[Zhen et~al.(2024)Zhen, Qiu, Chen, Yang, Yan, Du, Hong, and Gan]{zhen20243dvla}
Haoyu Zhen, Xiaowen Qiu, Peihao Chen, Jincheng Yang, Xin Yan, Yilun Du, Yining Hong, and Chuang Gan.
\newblock 3d-{VLA}: A 3d vision-language-action generative world model.
\newblock In \emph{Forty-first International Conference on Machine Learning, {ICML}}, 2024.

\bibitem[Zhu et~al.(2023)Zhu, Yang, Wu, Huang, Zhang, He, He, Zhao, Shen, Qiao, and Ouyang]{zhu2023ponderv2}
Haoyi Zhu, Honghui Yang, Xiaoyang Wu, Di~Huang, Sha Zhang, Xianglong He, Tong He, Hengshuang Zhao, Chunhua Shen, Yu~Qiao, and Wanli Ouyang.
\newblock Ponderv2: Pave the way for 3d foundation model with a universal pre-training paradigm.
\newblock In \emph{arXiv preprint arXiv:2310.08586}, 2023.

\bibitem[Zhu et~al.(2024)Zhu, Wang, Huang, Ye, Ouyang, and He]{zhu2024pointcloudmatters}
Haoyi Zhu, Yating Wang, Di~Huang, Weicai Ye, Wanli Ouyang, and Tong He.
\newblock Point cloud matters: Rethinking the impact of different observation spaces on robot learning.
\newblock In \emph{The Thirty-eight Conference on Neural Information Processing Systems Datasets and Benchmarks Track, {NIPS}}, 2024.

\bibitem[Zhu et~al.(2025)Zhu, Yang, Wang, Yang, Wang, and He]{zhu2025spa}
Haoyi Zhu, Honghui Yang, Yating Wang, Jiange Yang, Limin Wang, and Tong He.
\newblock {SPA}: 3d spatial-awareness enables effective embodied representation.
\newblock In \emph{The Thirteenth International Conference on Learning Representations, {ICLR}}, 2025.

\end{thebibliography}
\bibliographystyle{plainnat}

%%%%%%%%%%%%%%%%%%%%%%%%%%%%%%%%%%%%%%%%%%%%%%%%%%%%%%%%%%%%
\newpage

\appendix

\section{Details of Experimental Setup}

\subsection{Policy Network}\label{appendix:policy_network}

To evaluate how well Vision Foundation Models (VFMs) support advanced Vision-Language Action (VLA) models, we use the RDT \citep{liu2025rdt} architecture as our evaluation policy network, which has demonstrated excellent scalability and strong performance in diffusion-based policy learning. Diffusion timestamps and robot kinematic information are integrated into the policy network using AdaLN-Zero \citep{Peebles2022DiT}. The vision and language embeddings are used as the Keys and Values in the cross-attention layers to be integrated into the policy network alternately \citep{liu2025rdt}. The Transformer architecture has a hidden dimension of 384, with 6 attention heads, and 12 layers.

For action generation, we use a flow-matching model similar to \citep{black2024pi0}. Diffusion timestamps are treated as continuous values within the range $[0, 1]$; we do not discretize them. Instead, they are represented using a Fourier embedding with a scale of 0.2 \citep{dong2024cleandiffuser}. During training, diffusion timestamps are sampled from a uniform distribution over the interval $[0, 1]$. For inference, we solve the corresponding ODE using the Euler method, dividing the interval $[0, 1]$ into equal-sized steps.

\subsection{Details of Benchmarks}\label{appendix:benchmarks}

\begin{figure}[ht]
    \centering
    \includegraphics[width=1.0\linewidth]{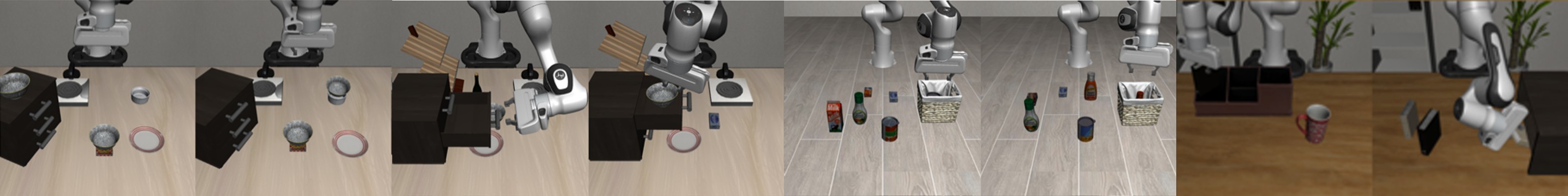}
    \vspace{-10pt}
    \caption{\small \textbf{LIBERO simulation benchmark.} We conduct experiments on 40 tasks from four task suites in the LIBERO benchmark. We show two task examples for each suite here.}
    \label{fig:libero_task}
\end{figure}

\begin{figure}[ht]
    \centering
    \includegraphics[width=1.0\linewidth]{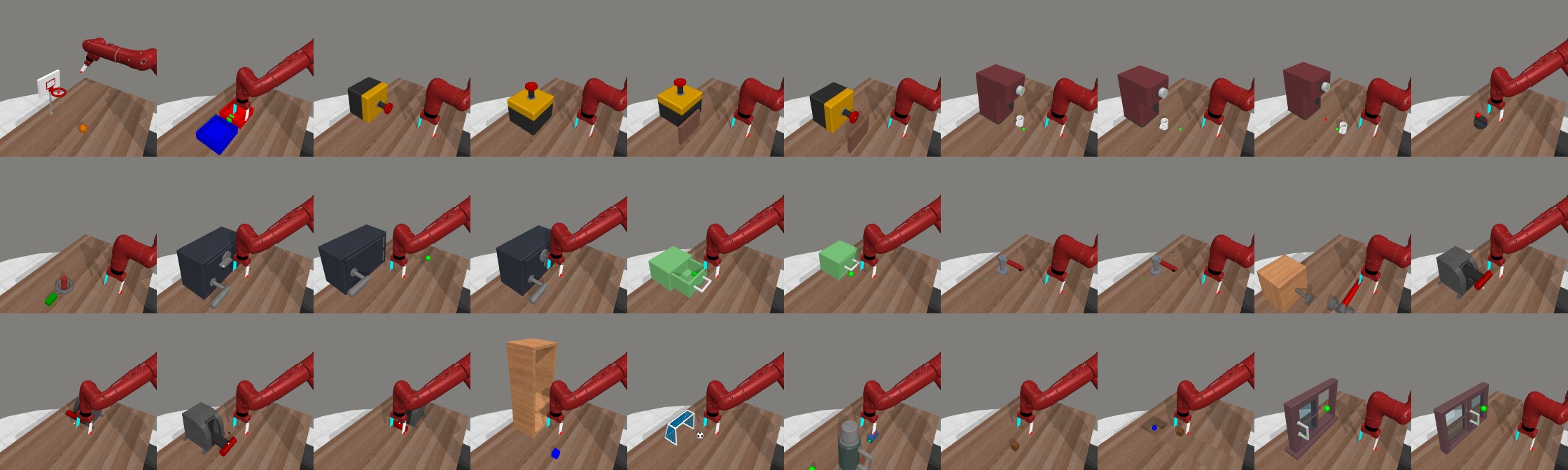}
    \caption{\small \textbf{MetaWorld simulation benchmark.} We conduct experiments on 30 tasks of three difficulty levels in the MetaWorld benchmark. We show all task examples here.}
    \label{fig:metaworld_task}
\end{figure}

\begin{figure}[ht]
    \centering
    \includegraphics[width=1.0\linewidth]{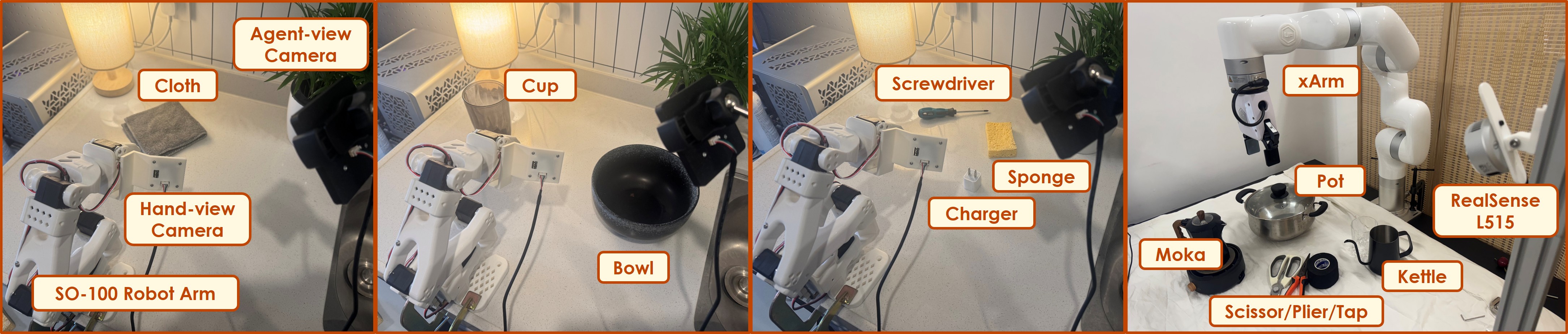}
    \vspace{-10pt}
    \caption{\small{\textbf{Real-world experimental setups.} We conduct experiments on both SO100 and xArm platform. For each robot, we design a suite of 10 tabletop tasks involving diverse objects.}}
    \label{fig:so100_task}
\end{figure}

\begin{table}[htbp]
\centering
\caption{{\small Task description of each task in the LIBERO benchmark.}}
\scalebox{0.8}{
\label{tab:libero_task}
\begin{tabular}{l|l}
\toprule
\multicolumn{1}{c|}{Task Suite}  & \multicolumn{1}{c}{Task Description}                                                       \\
\midrule
\multirow{10}{*}{LIBERO-Goal}    & \small open the middle layer of the drawer                                                        \\
                                 & \small put the bowl on the stove                                                                  \\
                                 & \small put the wine bottle on the top of the drawer                                               \\
                                 & \small open the top layer of the drawer and put the bowl inside                                   \\
                                 & \small put the bowl on the top of the drawer                                                      \\
                                 & \small push the plate to the front of the stove                                                   \\
                                 & \small put the cream cheese on the bowl                                                           \\
                                 & \small turn on the stove                                                                          \\
                                 & \small put the bowl on the plate                                                                  \\
                                 & \small put the wine bottle on the rack                                                            \\
\midrule
\multirow{10}{*}{LIBERO-Spatial} & \small pick the akita black bowl between the plate and the ramekin and place it on the plate      \\
                                 & \small pick the akita black bowl next to the ramekin and place it on the plate                    \\
                                 & \small pick the akita black bowl from table center and place it on the plate                      \\
                                 & \small pick the akita black bowl on the cookies box and place it on the plate                     \\
                                 & \small pick the akita black bowl in the top layer of the wooden cabinet and place it on the plate \\
                                 & \small pick the akita black bowl on the ramekin and place it on the plate                         \\
                                 & \small pick the akita black bowl next to the cookies box and place it on the plate                \\
                                 & \small pick the akita black bowl on the stove and place it on the plate                           \\
                                 & \small pick the akita black bowl next to the plate and place it on the plate                      \\
                                 & \small pick the akita black bowl on the wooden cabinet and place it on the plate                  \\
\midrule
\multirow{10}{*}{LIBERO-Object}  & \small pick the alphabet soup and place it in the basket                                          \\
                                 & \small pick the cream cheese and place it in the basket                                           \\
                                 & \small pick the salad dressing and place it in the basket                                         \\
                                 & \small pick the bbq sauce and place it in the basket                                              \\
                                 & \small pick the ketchup and place it in the basket                                                \\
                                 & \small pick the tomato sauce and place it in the basket                                           \\
                                 & \small pick the butter and place it in the basket                                                 \\
                                 & \small pick the milk and place it in the basket                                                   \\
                                 & \small pick the chocolate pudding and place it in the basket                                      \\
                                 & \small pick the orange juice and place it in the basket                                           \\
\midrule
\multirow{10}{*}{LIBERO-Long}    & \small put both the alphabet soup and the tomato sauce in the basket                              \\
                                 & \small put both the cream cheese box and the butter in the basket                                 \\
                                 & \small turn on the stove and put the moka pot on it                                               \\
                                 & \small put the black bowl in the bottom drawer of the cabinet and close it                        \\
                                 & \small put the white mug on the left plate and put the yellow and white mug on the right plate    \\
                                 & \small pick up the book and place it in the back compartment of the caddy                         \\
                                 & \small put the white mug on the plate and put the chocolate pudding to the right of the plate     \\
                                 & \small put both the alphabet soup and the cream cheese box in the basket                          \\
                                 & \small put both moka pots on the stove                                                            \\
                                 & \small put the yellow and white mug in the microwave and close it                                 \\
\bottomrule
\end{tabular}}
\end{table}

\begin{table}[htbp]
  \centering
  \caption{\small Task description of each task in the MetaWorld benchmark.}
  \label{tab:metaworld_task}
  \scalebox{0.8}{
  \begin{tabular}{l|l}
    \toprule
    \textbf{Task Name} & \textbf{Task Description} \\
    \midrule
    basketball & Dunk the basketball into the basket. \\
    bin-picking & Grasp the puck from one bin and place it into another bin. \\
    button-press & Press a button. \\
    button-press-topdown & Press a button from the top. \\
    button-press-topdown-wall & Bypass a wall and press a button from the top. \\
    button-press-wall & Bypass a wall and press a button. \\
    coffee-button & Push a button on the coffee machine. \\
    coffee-pull & Pull a mug from a coffee machine. \\
    coffee-push & Push a mug under a coffee machine. \\
    dial-turn & Rotate a dial 180 degrees. \\
    disassemble & Pick a nut out of the peg. \\
    door-lock & Lock the door by rotating the lock clockwise. \\
    door-open & Open a door with a revolving joint. \\
    door-unlock & Unlock the door by rotating the lock counter-clockwise. \\
    drawer-close & Push and close a drawer. \\
    drawer-open & Open a drawer. \\
    faucet-close & Rotate the faucet clockwise. \\
    faucet-open & Rotate the faucet counter-clockwise. \\
    hammer & Hammer a screw on the wall. \\
    handle-press & Press a handle down. \\
    handle-press-side & Press a handle down sideways. \\
    handle-pull & Pull a handle up. \\
    handle-pull-side & Pull a handle up sideways. \\
    shelf-place & Pick and place a puck onto a shelf. \\
    soccer & Kick a soccer into the goal. \\
    stick-push & Grasp a stick and push a box using the stick. \\
    sweep & Sweep a puck off the table. \\
    sweep-into & Sweep a puck into a hole. \\
    window-close & Push and close a window. \\
    window-open & Push and open a window. \\
    \bottomrule
  \end{tabular}}
\end{table}

\begin{table}[htbp]
\centering
\caption{\small\textbf{Task description of each task in the SO100 and xArm benchmark.} As each parameter combination introduces one task, each task suite has 10 tasks in total. For each task, we test the model for 10 trials.}
\label{tab:so100_task}
\scalebox{0.7}{
\begin{tabular}{l|l|l}
\toprule
Task Suite & Task Description & Parameter \\
\midrule

\multirow{4}{*}{SO100} & 
\small pick \texttt{[A]} and place it on the \texttt{[B]} side of the table & 
\small \texttt{[A]: ["screwdriver", "sponge", "charger"]}, \texttt{[B]: ["left", "right"]} \\
 & 
\small move \texttt{[A]} to the center of the table &
\small \texttt{[A]: ["cup", "bowl"]} \\
& 
\small pick the cloth and wipe the table & 
\small \texttt{None} \\
& 
\small unfold the cloth & 
\small \texttt{None} \\

\midrule

\multirow{4}{*}{xArm} &

\small pick \texttt{[A]} and place it on the \texttt{[B]} side of the table & 
\small \texttt{[A]: ["scissor", "plier", "tap"]}, \texttt{[B]: ["left", "right"]} \\
 & 
\small open the pot lid \texttt{or} put the lid on the pot &
\small \texttt{[open, close]} \\
& 
\small pour the water from the kettle into the cup & 
\small \texttt{None} \\
& 
\small place the Moka pot on the cooker & 
\small \texttt{None} \\

\bottomrule
\end{tabular}}
\end{table}

\textbf{LIBERO.} The LIBERO simulation benchmark \citep{liu2023libero} features a Franka Emika Panda arm in simulation across four challenging task suites: \textit{Goal}, \textit{Spatial}, \textit{Object}, and \textit{Long}. Each suite comprises 10 tasks with 500 demonstrations and is designed to investigate controlled knowledge transfer related to goal variations, spatial configurations, object types, and long-horizon tasks. Unlike prior work \citep{kim2024openvla, kim2025openvlaoft}, we do not filter out unsuccessful demonstrations, aiming for a more realistic evaluation setting. For policy training, the model predicts action chunks of length 16; after each chunk prediction, 8 steps are executed before generating the next chunk. The observation space includes 2-view RGB images at the current time step, without historical observations. During evaluation, following \citet{liu2023libero}, each task is tested over 50 trials with 3 different random seeds, and success rates are reported. To provide a clearer understanding of the task suites, we present agent-view observations in \Cref{fig:libero_task} and detailed task descriptions in \Cref{tab:libero_task}.

\textbf{MetaWorld.} The MetaWorld simulation benchmark \citep{yu2019metaworld} includes 50 distinct tabletop manipulation tasks using a Sawyer robot arm. We select 30 tasks from \textit{easy}, \textit{medium}, and \textit{very hard} difficulty levels to evaluate VLA models. We use a scripted policy to collect 20 demonstrations for each task. For policy training, the model predicts action chunks of length 16; after each chunk prediction, 16 steps are executed before generating the next chunk. The observation space consists of a single RGB image at the current time step, without historical observations. During evaluation, each task is tested over 50 trials with 3 different random seeds, and success rates are reported. To better illustrate the task suites, we show agent-view observations in \Cref{fig:metaworld_task} and task descriptions in \Cref{tab:metaworld_task}.

\textbf{SO100 Robot Manipulation.} The SO100 robot \citep{cadene2024lerobot} is a low-cost, open-source 6-DoF manipulator, with both the leader and follower arms costing approximately \$250. We assemble the hardware using a 3D-printed kit provided by the open-source community. The robot has two RGB cameras: one mounted on the wrist and the other positioned to provide a third-person view. Both cameras operate at a resolution of 640×480 and 25 FPS. The robot controller runs at 30Hz, and actions are defined as target absolute joint angles. Due to its low-cost design, the platform has several hardware limitations, including significant arm jitter, low load capacity, and occasional camera lag, which present practical challenges for developing embodied AI systems. However, given the increasing adoption of such affordable open-source robots by the research community, we believe that evaluating models on these lower-performance systems offers valuable insights and broader applicability. We design four categories of tabletop manipulation tasks for the SO100 setup: \textit{\textbf{(1) Pick\&Place:}} involving 3 objects and 2 placement zones (6 tasks), \textit{\textbf{(2) MoveTo:}} navigating 2 objects to a single target zone (2 tasks), \textit{\textbf{(3) Wipe:}} picking up a cloth and wiping the table (1 task), and \textit{\textbf{(4) Unfold:}} unfolding a cloth (1 task). In total, we evaluate performance on 10 distinct tasks. Language instructions for each task are listed in \Cref{tab:so100_task}, and visual examples of the task environments are shown in \Cref{fig:so100_task}.

During data collection, we record 20 demonstrations per task. For policy training, the model predicts an action chunk of length 64; after each chunk prediction, 40 steps are executed before generating the next chunk. The observation space includes 2-view RGB images at the current time step, along with the absolute joint angles from the current and previous 10 steps. During evaluation, each task is tested over 10 trials, and success rates are reported.

\textbf{xArm Robot Manipulation.} xArm is a high-performance 7-DoF manipulator. The robot is equipped with a third-person view Intel RealSense L515 LiDAR camera, operating at 640$\times$480 resolution and 30 FPS. We collect both RGB and depth images from the camera. The robot controller runs at 30Hz, and actions are defined as target absolute joint angles. We design four categories of tabletop manipulation tasks for the xArm setup: \textit{\textbf{(1) Pick\&Place:}} involving 3 objects and 2 placement zones (6 tasks), \textit{\textbf{(2) Pot:}} taking off or putting on the pot lid (2 tasks), \textit{\textbf{(3) Pour:}} pouring water from the kettle into the cup (1 task), and \textit{\textbf{(4) Moka:}} placing the Moka pot on the cooker (1 task). In total, we evaluate performance on 10 distinct tasks. Language instructions for each task are listed in \Cref{tab:so100_task}, and visual examples of task environments are shown in \Cref{fig:so100_task}.

During data collection, we record 20 demonstrations per task. For policy training, the model predicts an action chunk of length 64; after each chunk prediction, 40 steps are executed before generating the next chunk. The observation space includes a third-person view RGB image at the current time step, as well as the absolute joint angles from the current and previous 10 steps. During evaluation, each task is tested over 10 trials, and success rates are reported.

\begin{table}[h]
\centering

\caption{\small\textbf{Hyperparameters for EmbodiedMAE training.} Since we use pre-training and distillation for different model scales, we use \textbf{(P)} to denote pre-training hyperparameters and \textbf{(D)} to denote distillation hyperparameters.}
\scalebox{0.9}{
\label{tab:hyperparams}
\begin{tabular}{l|l}
\toprule
Hyperparameters      & Values                                                                        \\
\midrule
GPUs                 & \small 8xNVIDIA L40 (60GB) \textbf{(P)} or 4xNVIDIA Geforce RTX4090 (24GB) \textbf{(D)}                \\
learning rate        & \small 3e-4 peak LR (500 steps linear warmup, 300k steps cosine decay to 3e-6)       \\
batch size           & \small 512                                                                           \\
training steps       & \small 200K \textbf{(P)} 100K \textbf{(D)}                                                             \\
input modalities     & \small 224x224x3 RGB images, 224x224 Depth maps, 8,192 Point Clouds                   \\
image augmentations  & \small ColorJitter(brightness=0.1, contrast=0.1, saturation=0.1, hue=0.05)           \\
trainable parameters & \small 1.1B Giant \textbf{(P)} 304M Large, 87M Base, 22M Small \textbf{(D)} encoders, and 44M decoders \\
mask ratio           & \small 84\% \textbf{(P)} 90\% \textbf{(D)}                                                             \\
Distillation $\beta$ & 1.0 \\
\bottomrule
\end{tabular}}
\end{table}

\subsection{DINOv2-RGBD Baseline}
\label{appendix:dino_rgbd}

To establish a reliable and effective RGBD baseline for practical applications, we follow the approach outlined in \citet{zhu2024pointcloudmatters}, designing a method that naively incorporates depth information based on DINOv2. We introduce an additional Conv2D layer to patchify the depth map, summing the resulting patches with DINOv2's RGB patchifying output before encoding through the DINOv2 Encoder. We initialize the depth patchifier's weights and biases to zero, ensuring that the representation model remains functionally equivalent to DINOv2 at the beginning of training. During training, we update only the depth patchifier's gradients, allowing depth information to be learned following DINOv2's prior knowledge.

%%%%%%%%%%%%%%%%%%%%%%%%%%%%%%%%%%%%%%%%%%%%%%%%%%%%%%%%%%%%

\newpage

\section*{NeurIPS Paper Checklist}

\begin{enumerate}

\item {\bf Claims}
    \item[] Question: Do the main claims made in the abstract and introduction accurately reflect the paper's contributions and scope?
    \item[] Answer: \answerYes{} % Replace by \answerYes{}, \answerNo{}, or \answerNA{}.
    \item[] Justification: The main claims made in the abstract and introduction accurately reflect the paper's contributions and scope.

\item {\bf Limitations}
    \item[] Question: Does the paper discuss the limitations of the work performed by the authors?
    \item[] Answer: \answerYes{} % Replace by \answerYes{}, \answerNo{}, or \answerNA{}.
    \item[] Justification: The paper have discussed the limitations of the work.

\item {\bf Theory assumptions and proofs}
    \item[] Question: For each theoretical result, does the paper provide the full set of assumptions and a complete (and correct) proof?
    \item[] Answer: \answerNA{} % Replace by \answerYes{}, \answerNo{}, or \answerNA{}.
    \item[] Justification: The paper does not include theoretical results.

    \item {\bf Experimental result reproducibility}
    \item[] Question: Does the paper fully disclose all the information needed to reproduce the main experimental results of the paper to the extent that it affects the main claims and/or conclusions of the paper (regardless of whether the code and data are provided or not)?
    \item[] Answer: \answerYes{} % Replace by \answerYes{}, \answerNo{}, or \answerNA{}.
    \item[] Justification: We provide every detail to reproduce the main experimental results.

\item {\bf Open access to data and code}
    \item[] Question: Does the paper provide open access to the data and code, with sufficient instructions to faithfully reproduce the main experimental results, as described in supplemental material?
    \item[] Answer: \answerYes{} % Replace by \answerYes{}, \answerNo{}, or \answerNA{}.
    \item[] Justification: We will open-source the code in a few days after submission.

\item {\bf Experimental setting/details}
    \item[] Question: Does the paper specify all the training and test details (e.g., data splits, hyperparameters, how they were chosen, type of optimizer, etc.) necessary to understand the results?
    \item[] Answer: \answerYes{} % Replace by \answerYes{}, \answerNo{}, or \answerNA{}.
    \item[] Justification:  We provide every detail in training and testing.

\item {\bf Experiment statistical significance}
    \item[] Question: Does the paper report error bars suitably and correctly defined or other appropriate information about the statistical significance of the experiments?
    \item[] Answer: \answerYes{} % Replace by \answerYes{}, \answerNo{}, or \answerNA{}.
    \item[] Justification: Yes. We report error bars in the learning curve.

\item {\bf Experiments compute resources}
    \item[] Question: For each experiment, does the paper provide sufficient information on the computer resources (type of compute workers, memory, time of execution) needed to reproduce the experiments?
    \item[] Answer: \answerYes{} % Replace by \answerYes{}, \answerNo{}, or \answerNA{}.
    \item[] Justification: Yes. We list the GPU and CPU resources.
    
\item {\bf Code of ethics}
    \item[] Question: Does the research conducted in the paper conform, in every respect, with the NeurIPS Code of Ethics \url{https://neurips.cc/public/EthicsGuidelines}?
    \item[] Answer: \answerYes{} % Replace by \answerYes{}, \answerNo{}, or \answerNA{}.
    \item[] Justification: Yes, we do.

\item {\bf Broader impacts}
    \item[] Question: Does the paper discuss both potential positive societal impacts and negative societal impacts of the work performed?
    \item[] Answer: \answerYes{} % Replace by \answerYes{}, \answerNo{}, or \answerNA{}.
    \item[] Justification: Yes, we do.
    
\item {\bf Safeguards}
    \item[] Question: Does the paper describe safeguards that have been put in place for responsible release of data or models that have a high risk for misuse (e.g., pretrained language models, image generators, or scraped datasets)?
    \item[] Answer: \answerNA{} % Replace by \answerYes{}, \answerNo{}, or \answerNA{}.
    \item[] Justification: The paper poses no such risks.

\item {\bf Licenses for existing assets}
    \item[] Question: Are the creators or original owners of assets (e.g., code, data, models), used in the paper, properly credited and are the license and terms of use explicitly mentioned and properly respected?
    \item[] Answer: \answerYes{} % Replace by \answerYes{}, \answerNo{}, or \answerNA{}.
    \item[] Justification: Yes, they are.

\item {\bf New assets}
    \item[] Question: Are new assets introduced in the paper well documented and is the documentation provided alongside the assets?
    \item[] Answer: \answerYes{} % Replace by \answerYes{}, \answerNo{}, or \answerNA{}.
    \item[] Justification: Yes.

\item {\bf Crowdsourcing and research with human subjects}
    \item[] Question: For crowdsourcing experiments and research with human subjects, does the paper include the full text of instructions given to participants and screenshots, if applicable, as well as details about compensation (if any)? 
    \item[] Answer: \answerNA{} % Replace by \answerYes{}, \answerNo{}, or \answerNA{}.
    \item[] Justification: We do not include crowdsourcing experiments or research with human subjects.

\item {\bf Institutional review board (IRB) approvals or equivalent for research with human subjects}
    \item[] Question: Does the paper describe potential risks incurred by study participants, whether such risks were disclosed to the subjects, and whether Institutional Review Board (IRB) approvals (or an equivalent approval/review based on the requirements of your country or institution) were obtained?
    \item[] Answer: \answerNA{} % Replace by \answerYes{}, \answerNo{}, or \answerNA{}.
    \item[] Justification: We do not include crowdsourcing experiments or research with human subjects.

\item {\bf Declaration of LLM usage}
    \item[] Question: Does the paper describe the usage of LLMs if it is an important, original, or non-standard component of the core methods in this research? Note that if the LLM is used only for writing, editing, or formatting purposes and does not impact the core methodology, scientific rigorousness, or originality of the research, declaration is not required.
    %this research? 
    \item[] Answer: \answerNA{} % Replace by \answerYes{}, \answerNo{}, or \answerNA{}.
    \item[] Justification: The core method development in this research does not involve LLMs.

\end{enumerate}

\end{document}